\documentclass[journal]{IEEEtran}
\ifCLASSINFOpdf
\else
\fi
\usepackage{graphicx}
\usepackage{caption,tabularx,booktabs}
\usepackage{cite}
\usepackage{amsmath,amssymb,amsfonts}
\usepackage{algorithm}
\usepackage{algorithmic}
\usepackage{textcomp}
\usepackage[pagebackref=true,breaklinks=true,letterpaper=true,colorlinks,bookmarks=false]{hyperref}
\usepackage{multirow}
\usepackage{graphicx}
\usepackage[table,xcdraw]{xcolor}
\usepackage{ifthen, xcolor}
\usepackage{etoolbox}

\newboolean{cs}
\setboolean{cs}{false}
\newcommand{\setcstrue}{\textcolor{blue}}
\newcommand{\setcsfalse}{}
\newcommand{\cs}{\ifthenelse{\boolean{cs}}{\setcstrue}{\setcsfalse}}

\def\BibTeX{{\rm B\kern-.05em{\sc i\kern-.025em b}\kern-.08em
    T\kern-.1667em\lower.7ex\hbox{E}\kern-.125emX}}

\usepackage{xparse,xcolor}

\begin{document}
%
% paper title
% Titles are generally capitalized except for words such as a, an, and, as,
% at, but, by, for, in, nor, of, on, or, the, to and up, which are usually
% not capitalized unless they are the first or last word of the title.
% Linebreaks \\ can be used within to get better formatting as desired.
% Do not put math or special symbols in the title.
\title{Orthogonal Deep Models As Defense Against Black-Box Attacks}
%
%
% author names and IEEE memberships
% note positions of commas and nonbreaking spaces ( ~ ) LaTeX will not break
% a structure at a ~ so this keeps an author's name from being broken across
% two lines.
% use \thanks{} to gain access to the first footnote area
% a separate \thanks must be used for each paragraph as LaTeX2e's \thanks
% was not built to handle multiple paragraphs
%

%
% \author{\uppercase{Mohammad A. A. K. Jalwana}\authorrefmark{1},
% \uppercase{\authorrefmark{2}, }\authorrefmark{3} AND \uppercase{}\authorrefmark{4}}
% \address[1-4]{Computer Science and Software Engineering, The University of Western Australia.}

\author{Mohammad A. A. K. Jalwana,
        Naveed Akhtar,
        Mohammed Bennamoun, 
        and~Ajmal Mian,\\% <-this % stops a space
        Computer Science and Software Engineering, The University of Western Australia.\\
        {\{mohammad.jalwana@research., naveed.akhtar@, mohammed.bennamoun@, ajmal.mian@\}uwa.edu.au}
}

% note the % following the last \IEEEmembership and also \thanks - 
% these prevent an unwanted space from occurring between the last author name
% and the end of the author line. i.e., if you had this:
% 
% \author{....lastname \thanks{...} \thanks{...} }
%                     ^------------^------------^----Do not want these spaces!
%
% a space would be appended to the last name and could cause every name on that
% line to be shifted left slightly. This is one of those "LaTeX things". For
% instance, "\textbf{A} \textbf{B}" will typeset as "A B" not "AB". To get
% "AB" then you have to do: "\textbf{A}\textbf{B}"
% \thanks is no different in this regard, so shield the last } of each \thanks
% that ends a line with a % and do not let a space in before the next \thanks.
% Spaces after \IEEEmembership other than the last one are OK (and needed) as
% you are supposed to have spaces between the names. For what it is worth,
% this is a minor point as most people would not even notice if the said evil
% space somehow managed to creep in.

% The paper headers
\markboth{Journal of \LaTeX\ Class Files,~Vol.~14, No.~8, August~2015}%
{Shell \MakeLowercase{\textit{et al.}}: Bare Demo of IEEEtran.cls for IEEE Journals}
% The only time the second header will appear is for the odd numbered pages
% after the title page when using the twoside option.
% 
% *** Note that you probably will NOT want to include the author's ***
% *** name in the headers of peer review papers.                   ***
% You can use \ifCLASSOPTIONpeerreview for conditional compilation here if
% you desire.

% If you want to put a publisher's ID mark on the page you can do it like
% this:
%\IEEEpubid{0000--0000/00\$00.00~\copyright~2015 IEEE}
% Remember, if you use this you must call \IEEEpubidadjcol in the second
% column for its text to clear the IEEEpubid mark.

% use for special paper notices
%\IEEEspecialpapernotice{(Invited Paper)}

% make the title area
\maketitle

% As a general rule, do not put math, special symbols or citations
% in the abstract or keywords.
\begin{abstract}
Deep learning has demonstrated  state-of-the-art performance for a variety of challenging computer vision tasks. 
On one hand, this has enabled deep visual models to pave the way for a plethora of critical applications like disease prognostics and smart surveillance.
On the other, deep learning has also been found vulnerable to adversarial attacks, which calls for new techniques to defend deep models against these attacks. 
%on the other side, vulnerability of deep model to low power adversarial signals has recently raised serious concerns. 
% However, on the darker side, deep models are not secure and are easily bamboozled by low power additive signals. 
%demands the need of reliable-models that can perform well in presence of common real-life intrusions.  
%JalwanaIEEEReviewChangeStart
\cs{Among the attack algorithms, the black-box schemes are of serious practical concern since they only need  publicly available knowledge of the targeted model.}
 We carefully analyze the inherent weakness of deep models in black-box settings where the attacker may develop the attack using a model similar to the targeted model. 
% Below line was part earlier
% We carefully analyze the inherent weakness of deep models in black-box settings where the attacker may develop the attack using a model similar to the targeted model.
%JalwanaIEEEReviewChangeStop
%over a simplified setup where models of similar architectures are considered.
Based on our analysis, we introduce a novel gradient regularization scheme that encourages the internal representation of a deep model to be orthogonal to another, even if the architectures of the two models are  similar. 
Our unique constraint allows a model  to concomitantly endeavour for higher accuracy while maintaining   near orthogonal alignment of gradients with respect to a reference model. 
Detailed empirical study verifies that controlled misalignment of gradients under our orthogonality  objective significantly boosts a  model's robustness against  transferable \cs{black-box} adversarial attacks. 
In comparison to regular models, the orthogonal models are significantly more robust to a range of  $l_p$ norm bounded perturbations. 
%The gain in immunity does not compromises the general classification accuracy. 
%Our simple, yet powerful methodology can therefore benefit a range of applications.
We verify the effectiveness of our technique on a variety of large-scale models.

% We introduce a novel gradient regularization scheme that enables deep visual models to be orthogonal to others. 
% The unique constraint allows a model to concomitantly endeavour for higher accuracy with orthogonal alignment of gradients to a reference model. 
% Detailed empirical study verifies that controlled misalignment of gradients, under the objective of orthogonality, boosts model's robustness against the transferable adversarial attacks. 
% In comparison to ordinary models, the orthogonal models are significantly more robust for a range of semi/imperceptible perturbations defined by the $l_p$ norm. 
% The gain in immunity is does not compromises the general classification accuracy. 
% Our simple and yet powerful methodology can therefore provide security to a range of computer vision applications.
\end{abstract}

% Note that keywords are not normally used for peerreview papers.

\section{Introduction}
\label{sec:introduction}
Deep learning has enabled machines to achieve human level performance in numerous computer vision tasks, including image classification \cite{krizhevsky2012imagenet, simonyan2014very}, object detection~\cite{redmon2017yolo9000, ren2015faster}, semantic segmentation \cite{long2015fully, chen2018encoder} and image captioning~\cite{wu2017image}.
% However, despite their impressive accuracy, they are found to be vulnerable to imperceptible additive signals known as adversarial perturbations. 
However, despite their impressive accuracy, they are vulnerable to adversarial inputs~\cite{szegedy2013intriguing}. 
These inputs - a.k.a.~adversarial examples - are crafted by a careful manipulation of the original input signals.
The resulting adversarial examples  appear natural to humans but can completely alter the output of a deep model. 
This intriguing brittleness of deep learning is currently being actively investigated by the research community~\cite{akhtar2018threat}. 

Recent years have seen numerous algorithms to compute adversarial  perturbations to inputs that can  fool deep models.
These techniques can be broadly categorized into gradient-based and gradient-free schemes \cite{akhtar2018threat}. 
Generally, the gradient-based approaches accumulate the raw gradients of a model w.r.t.~input via iterative traversal of the model's loss surface.
These gradients are subsequently used for targeted or non-targeted fooling of the model.
Gradient-free techniques craft adversarial examples in  a local gradient agnostic manner, e.g.,~by iteratively taking feedback from model predictions and exploiting it for model fooling~\cite{andriushchenko2019square, chen2019hopskipjumpattack, ilyas2018black}.
It is claimed that gradient-based techniques pose a serious threat to deep learning \cite{madry2017towards}.  

\begin{figure*}[t]
    \centering
    \includegraphics[width= \textwidth]{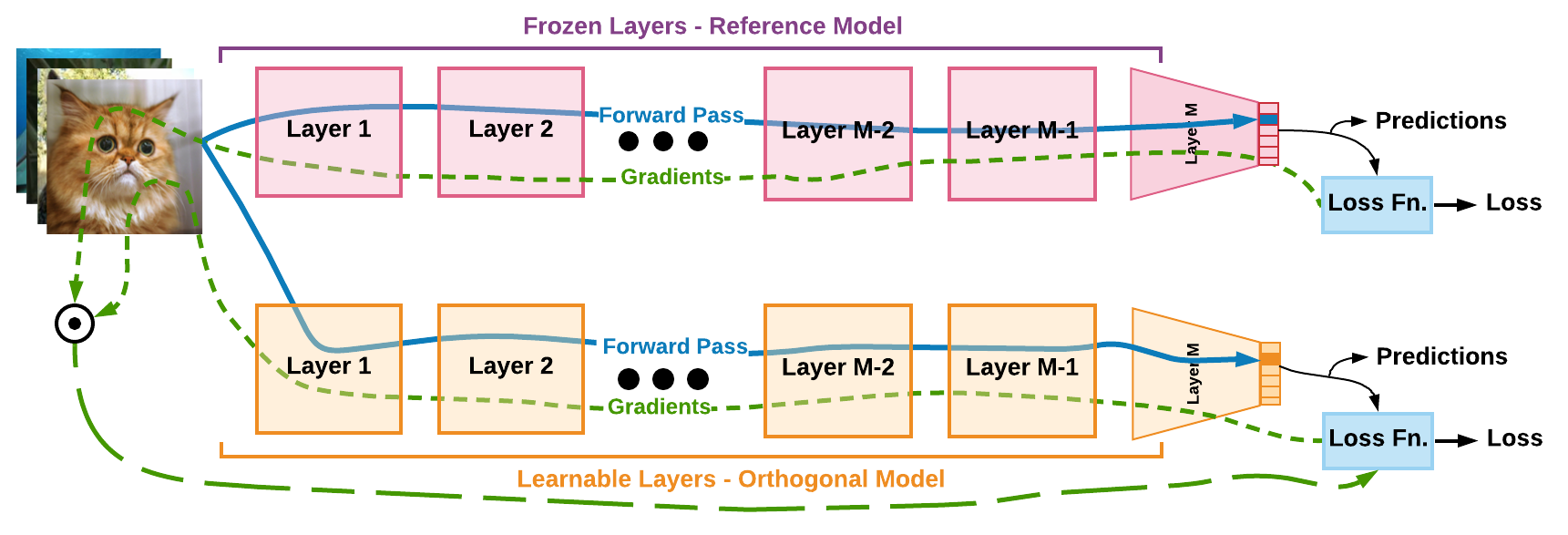}
    \caption{Schematics of our framework to train orthogonal models.  On the top, a trained reference model with frozen parameters is used for the computation of reference gradients. The model to be trained, shown at the bottom, exploits the reference gradients to align its gradients in an orthogonal manner. Control over the direction of  gradients is explicitly enabled through our novel similarity loss objective.}
    \label{fig:architecture}
\end{figure*}

Gradient-based perturbations exhibit high fooling rates under white-box setting, where complete knowledge of the targeted model is available. 
Surprisingly, these perturbations also transfer well to models under the more pragmatic black-box setting - where no knowledge of the target model is available. This phenomenon enables the attackers to craft malicious examples using a surrogate model \cite{liu2016delving}. % regardless of the target model. %Ajmal: not exactly regardless because there is explicit training of the surrogate model (or implicit assumption due to common training data) that the surrogate model follows the target model decision boundaries 
Such insidious nature of perturbations has raised serious concerns for the practical deployment of deep visual models in sensitive applications like self driving cars, medical diagnosis, face-recognition and many others.
%In most real life scenarios, knowledge of the model to be attacked is not available.  

The pivotal role of gradients in transferable adversarial examples naturally leads to the intuition of changing gradient relations among different models to induce immunity against attacks. 
Gradients of a model depend on its internal architectural layers that govern the flow of data from input to output. 
This inherent dependency of the gradients over the architecture allows the networks of varying architectures to have somewhat different gradients regardless of similar functional objective \cite{liu2016delving}.
However, different models with the same architecture may still have highly correlated gradients.

%---------------------

Previous studies \cite{liu2016delving, dong2018boosting, xie2019improving} demonstrate that perturbations generated from a model maximally transfer to models with similar architectures. 
We argue that the underlying similarity of gradients facilitate this phenomena. This intuition is based on the fundamental dependency of perturbations over model gradients, which allows the proximity in gradients of different models to manifest in their respective perturbations. 
Models are therefore, more susceptible to perturbations that are generated from a similar architecture. In this paper, we propose to mitigate the vulnerability of deep models to transferable (black-box) attacks by decreasing the correlation of  model gradients. 

For empirical investigation of the above arguments, we introduce a framework that allows explicit control over model gradients with respect to a reference model. 
Figure \ref{fig:architecture} illustrates the architectural view of our scheme.
It highlights the novel gradient based constraint that allows a given model (below) to strive for higher classification accuracy along with orthogonal alignment of its gradients to a reference model (top). In order to remove the disparity between the  gradients due to architectural differences, we keep the architectures similar and study the impact of gradient similarity over adversarial transferability. 
% In this work, to observe the standalone effects of gradient  disparity on attack transferability, we disentangle the architectural influence by varying the gradients of same architecture. Presented empirical study shows that randomly initialized models of the same architecture have linearly dependent gradients when trained on the same distribution. 
% Our systematic treatment of the problem, along with comprehensive experimentation demonstrates that the disparity in model's gradient space improves the network immunity against transferable attacks.
% As a side-affect, the models undergo little degradation in standard classification accuracy.
% The experiments empirically validate the enhanced immunity of  various models including VGG-11, ResNet-20 and ResNet-32 over the CIFAR-10 dataset. From practical point of view, the immunity over more challenging and widely used ImageNet ILSVRC2012 is also studied over the VGG-16 model. The results indicate that orthogonality generally helps in enhancement of robustness against the adversarial attacks.    

Our detailed analysis based on a systematic treatment of the problem demonstrates that the robustness of models against transferable attacks increases as the alignment between the model gradients decreases. %As a side-affect, standard classification accuracy of models suffers slightly.
Our experiments establish the enhanced immunity of a variety of models trained over CIFAR-10 and ImageNet ILSVRC 2012 datasets against transferable attacks using our technique. 
% IEEEReviewChangesJalwana
\cs{The enhanced robustness is significantly higher for the imperceptible range of perturbations. For instance, robustness of VGG-11 (CIFAR-10) model improves by $36.7\%$ against FGSM, $67.8\%$ against PGD, $66.8\%$ against I-FGSM and $55.8\%$ against the MI-FGSM generated perturbations with $\epsilon=0.05$.}
The major contributions of this work are summarized as:
\begin{itemize}
\item We propose a unique adversarial defense that is based on the orthogonality of deep models w.r.t.~a reference model.
\item We propose a novel gradient regularization scheme that enables a model to adjust the orientation of its gradients.
\item Our systematic empirical study analyzes the correlation between gradient similarity and adversarial transferability for independently trained models of similar architectures. It establishes that transferability of adversarial attacks between the models reduces as dissimilarity between their gradient increases.  
\end{itemize}

\section{Related Work}
Adversarial perturbations to inputs are primarily investigated in the related literature along the lines of attacking deep models and defending them against adversarial attacks~\cite{akhtar2018threat}. We first discuss the key contributions along these lines and then also describe a few recent attempts that go beyond the adversarial perspective of input perturbations. 

\subsection{Adversarial attacks}
Additive adversarial noise that can arbitrarily flip the prediction of a model made its first appearance in the seminal work of Szegedy et al.~\cite{szegedy2013intriguing}.
This work resulted in the development of numerous techniques to attack deep visual models.
Goodfellow et al.~\cite{goodfellow2014explaining} devised the Fast Gradient Sign Method (FGSM) to craft adversarial perturbations in a single gradient ascent step over the model's loss surface for the input.
Later,  Kurakin et al.~\cite{kurakin2016adversarial} advanced this scheme by introducing a multi-step version called Iterative FGSM (I-FGSM).
In continuation,  Dong et. al \cite{dong2018boosting} introduced momentum variables in the iterative traversal of loss surface and demonstrated enhanced transferability of adversarial perturbations.
Similarly, Xie et. al \cite{xie2019improving} further improved the I-FGSM and MI-FGSM by incorporation of random differentiable  transformations
like resizing and padding to the input image. Their technique is called Diverse Input I-FGSM (DI\textsuperscript{2}-FGSM). 
Further instances of the follow-up iterative algorithms are  Variance-Reduced I-FGSM (vr-IGSM)~\cite{wu2018understanding}  and PGD \cite{madry2017towards} etc. 

The above-mentioned algorithms and  other recent works \cite{moosavi2016deepfool, shi2019curls, rony2019decoupling,  croce2019sparse, zheng2019distributionally, ganeshan2019fda}  compute image-specific adversarial perturbations which appear as insignificant noise to the human eye but completely confuse  the models.
Moosavi-Dezfooli et al.~\cite{moosavi2017universal} first demonstrated the possibility of fooling deep models simultaneously on a large number of images with Universal Adversarial Perturbations. 
Later,  Akhtar et. al \cite{akhtar2019label} devised a label-universal technique to fool a model on an entire category of object in a targeted manner.  
%They demonstrated the practical importance of the attack over the safety critical face recognition and demonstrated that perturbation stay effective under quantization and realistic transformations of rotations and translation. 
It is now well-established that  adversarial examples computed by most techniques also transfer well across different deep learning models.
This insidious nature of adversarial perturbations is considered as a serious threat to practical deep learning~\cite{akhtar2018threat} and is fueling a very high level of research activity in this area.
%\textcolor{blue}{We will briefly discuss the major classification of adversarial attacks before providing overview of adversarial defense schemes.}

\subsubsection{Black-Box VS White-Box}
\cs{Adversarial attacks are broadly divided into white-box and black-box schemes \cite{akhtar2018threat}. This segregation is based on the amount of information available to the attacker to craft the perturbations.}

\cs{
In white-box attacks, the attacker assumes complete knowledge of the targeted model including its architecture, parameters values, training methods as well as the training data. The previously discussed schemes  \cite{goodfellow2014explaining, kurakin2016adversarial, dong2018boosting, xie2019improving, wu2018understanding, madry2017towards, moosavi2017universal} fall under this category. These techniques craft the perturbations based on the image and model specific gradients and, therefore, require full access to the model parameters as well as its architecture. Usually such attacks incur high fooling rates and are further differentiated by the number of iterations, choice of projection method and norm constraint to quantify the perturbation magnitude \cite{akhtar2018threat}.  
}

\cs{
In practice, intricate knowledge about the deployed model is seldom available. Therefore, 
attackers treat the target model as a black-box and utilize limited information made available in the public domain \cite{papernot2017practical}. These black-box attacks are broadly crafted either via  `query feedback' or `transferable attack' strategy \cite{bhambri2019survey}. In the query-based schemes, the attacker crafts the perturbations by repeatedly querying the targeted model \cite{ilyas2018black} and analysing the output. On the other hand, in transferable attack schemes, the attacker trains a substitute model to emulate the targeted model and then learns perturbations for the substitute model in a white-box setting. Such perturbations are known to fool models that are of different architectures and trained with different datasets~\cite{papernot2017practical, shi2019curls, dong2019evading}. Protection against the transferable schemes is considerably challenging as these attacks exploit similar generalization capability of models \cite{ilyas2019adversarial}.}

\subsection{Adversarial defenses}

A variety of techniques have also been proposed to counter the adversarial attacks~\cite{jia2019comdefend,  akhtar2018defense, liu2019detection}. 
These schemes aim to defend deep models against both image-specific~\cite{liu2019detection} as well as image-agnostic perturbations~\cite{akhtar2018defense}. 
Broadly, the defense schemes are developed along four major lines.

\vspace{1mm}
\noindent\textbf{Adversarial Training:} 
These schemes aim to boost the robustness of models by specialized augmentation in training dataset \cite{szegedy2013intriguing,goodfellow2014explaining,moosavi2016deepfool,akhtar2018threat, miyato2016adversarial,madry2017towards,woods2019reliable}. 
Adversarial examples are computed using the strongest available techniques and the model undergoes training on these examples along with the clean samples.  
Madry et. al \cite{madry2017towards} has systematically studied adversarial training over a variety of models and datasets.
Recently, they publicly released robustly trained models. 
Interestingly, beside significant performance degradation of these models, adversarial examples can still be computed for such models ~\cite{moosavi2016deepfool}. 

% It has been demonstrated that adversarial robustness downgrades the general classfication accuracy \cite{tsipras2019robustness} and there exists an inherent trade-off between them. 
% It is interesting to note that adversarial examples can still be computed for the adversarially trained models \cite{moosavi2016deepfool}. 

\vspace{0.5mm}
\noindent\textbf{Network Modifications:} 
Unlike adversarial training, these techniques rework some inner aspects of a model. 
This generally includes introduction of new layers and   enhancement of the loss function via regularization of the gradients.
For instance, auto-encoders have received special attention among the studies that introduce  additional layers.
Gu et al.~\cite{gu2014towards} and Bai et al.~\cite{bai2017alleviating} have  explored auto-encoders to mitigate the adversarial noise in the spatial domain.
% The role of auto-encoders layers to mitigate the adversarial noise was studied by Gu et. al\cite{gu2014towards}  and Bai et.al \cite{bai2017alleviating}. 
Similarly, \cite{gao2017deepcloak} introduced a masking layer to clean adversarial noise at the highest feature layer. 
% Besides the effectiveness towards the adversarial examples, the enhanced models can still be exploited to compute them. 
% Besides their effectiveness, demonstrated results suggest that the  resultant models remain prone to adversarial attacks.
Gradient regularization schemes \cite{ross2018improving,lyu2015unified,shaham2018understanding,woods2019reliable} are normally based on the observation that adversarial perturbations have smaller norms. 
Therefore, penalizing the degree of variation of output with respect to input can assist in detecting adversarial noise.

\vspace{1mm}
\noindent\textbf{Network Add-ons:}
External networks have also explored to aid in defense against adversarial attacks. 
Shen et. al \cite{shen2017ape} re-purposed the generative adversarial networks (GANs) to rectify the perturbed image. 
While Lee et. al \cite{lee2017generative} customized those to generate adversarial examples. 
The generated examples along with clean samples are utilized to train the ordinary models. 
The later setup has strong resemblance to adversarial training, however, the fixed presence and larger role of additional generator pushes this scheme in to this category \cite{akhtar2018threat}.
Similarly, Akhtar et al.~\cite{akhtar2018defense}  proposed a perturbation rectifying network (PRN).
This sub-network pre-processes the input images before passing it on to the classification model. 
Interestingly, training of PRN layers do not modify the weights of the  classification model and act as an effective defense against universal adversarial perturbations. 

\vspace{0.5mm}
\noindent\textbf{Input Transformations:}
Apart from the above-mentioned schemes, the line of work based on inherent brittleness of adversarial patterns has recently gained momentum. 
These schemes are based on the observation that perturbed inputs do not stay adversarial in the  presence of simple geometric transformations.  
Among the transformation-based defenses, Pixel deflection \cite{prakash2018deflecting}, BaRT \cite{raff2019barrage} and \cite{guo2017countering} are notable. 
An observation common among the mentioned studies is the general deterioration of classification accuracy in the presence of any defense strategy. 
% General classification accuracy over the clean images is usually compromised in almost all the defense strategies. 
% Among all the defense schemes, the general classification over the clean images is usually compromised. 
It has been demonstrated by  Carlini et al.~\cite{carlini2017adversarial, carlini2016defensive} and later by Athalye et al.~\cite{athalye2018robustness} that it is often possible to break the adversarial defenses by stronger adversarial attacks. 
Among all the defenses, adversarial training is considered to be the strongest.
However, it raises the computational budget considerably along with significant performance degradation  \cite{madry2017towards}. 

\subsection{Beyond attacks and defenses}  Recently, a handful of works \cite{santurkar2019computer,woods2019reliable,akhtar2019label} have demonstrated the usefulness of perturbations beyond the simple task of fooling.
In this aspect, \cite{santurkar2019computer} and \cite{woods2019reliable} showed that the perturbations generated for adversarially robust classifiers manifest the visual semantics of the target class. Interestingly, Santurkar et. al \cite{santurkar2019computer} cast a number of computer vision applications as adversarial attacks over the robust models. 
It includes image generation, inpainting, interactive image manipulation and image translation. 
However, the presented visual results are far from acceptable.
More recently, Jalwana et al.~\cite{ourcvpr2020}  demonstrated that attacks can also be a useful tool for model explanation.
However, the authors still advocate the need of defense techniques in adversarial settings. 
%Akhtar et. al \cite{akhtar2019label} developed a semi-universal attack scheme that demonstrated the existence of semantically meaningful patterns in perturbation for the ordinary classifiers. 
%The manifestation of geometric patterns provides insights towards understanding of the inner mechanics of deep visual models. It hints towards the semantical understanding of models \cite{akhtar2019label}. 
%Furthermore, adversarial attacks have also been utilized for inspection of classification regions as well as the boundaries \cite{akhtar2019label,moosavi2016deepfool}.

%-------------------------------------------------------------------------
% by reformulation its training objective

\subsection{Orthogonality In Deep Models}
\cs{Recently, the principle of orthogonality has been explored in the context of deep models. Zhang et al.~\cite{zhang2018cappronet} explored the orthogonal projections of features to devise a capsule projection network. Their work improves the classification accuracy of standard architectures over a variety of benchmark datasets. Similarly, \cite{gayer2020convolutional, jia2019orthogonal} have contributed towards the orthogonal regularization of model parameters. Jia et. al~\cite{jia2019orthogonal}  demonstrated that besides the enhanced accuracy, these models also have a superior natural resistance to common perturbations like blur, weather,  digitization etc.~without any explicit training. However, their work did not explore the robustness towards the engineered adversarial noises. In the next Section, we introduce and discuss our proposed technique for learning orthogonal models. Unlike existing works, our definition of orthogonality is based on the gradients rather than the parameters. }

\section{Proposed approach}
% I need to give a smoother entrance to this section

The prime objective of our technique is to misalign the gradients of a given model w.r.t a reference model. 
First, we introduce a metric that quantifies similarity between two models. 
We then reformulate the training objective %of ordinary models
to allow explicit control over the similarity.
% Before leading to detailed empirical evaluation, some qualitative aspects of gradient dissimilarity are inspected.  
Detailed discussion on the proposed algorithm is presented before analyzing the qualitative aspects of the gradient
disparity on model  decisions. 

We define the relative orientation of gradients as a measure of their similarity (`$\delta $'). 
Similarity between two normalized gradients  $\boldsymbol {g_1}$ and $\boldsymbol {g_2}$ is computed as the cosine of their mutual angle. For mathematical ease,  the multi-dimensional gradients are cast as $\mathbf{d}$ dimensional vectors.  The metric, as given in Equation \ref{dotProductEq}, has resemblance to identifying correlation between vectors.
\begin{align}
    \delta^i =  \boldsymbol{g}_1^i \odot \boldsymbol{g}_2^i  ~~~~ \text{s.t.} ~\boldsymbol{g}_1^i, \boldsymbol{g}_2^i \in \mathbb R^d,  
     \label{dotProductEq}
  \end{align}
where $\odot$ denotes vector dot product.  

We briefly discuss the commonly used training of ordinary models before introducing our enhanced loss function for orthogonal models. 
For a distribution $\mathcal{D}$ over images \textbf{$i \in \mathbb R^d$} with corresponding labels $l \in [K]$ and selected loss function $\mathcal L(\theta, i, l)$, the goal of ordinary training is to estimate the model parameters  $\theta$ that minimize the empirical risk
  \begin{align}
     \wp \!=\! \min_{{\theta}} \underset{{\boldsymbol{(i,l)}\sim \mathcal{D}}} {\mathbb E} \big[ \mathcal{L}(\theta, i, l)\big],
    \label{normalTrainingEq}
\end{align}
where $\mathbb E [.]$ is the Expectation operator. In this setting, gradients have complete liberty to follow any trajectory that leads to minimal risk or maximum accuracy. 
Nevertheless, with similar training data, architecture and hyper-parameter settings, gradients of different models induced by different initializations of parameters still have directional similarities.  
This can also be true for the models with slight architectural variations, which generally leads to good transferability of adversarial attacks across different models. 

We handle this discrepancy by allowing control over the gradient orientation of a model during its induction. To that end, we enhance the commonly used training objective with the help of the similarity metric defined in Equation~\ref{dotProductEq}. For a given reference model trained over the distribution $\mathcal{D}$, we train a misaligned model with parameters $\theta^*$ by  minimizing the empirical risk mentioned below
  \begin{align}
     \wp^* \!=\! \min_{\boldsymbol{\theta^*}} \underset{{\boldsymbol{(i,l)}\sim \mathcal{D}}} {\mathbb E} \big[ \mathcal{L}(\theta^*, i, l)  + \gamma\delta^i \big],
    \label{orthoTrainingEq}
\end{align}
where the hyper-parameter `$\gamma$'  enables explicit control over the gradient orientation during minimization.  As the value of this parameter increases, the gradients tend towards becoming orthogonal to those of the reference model with minimal loss of accuracy. %\textcolor{blue}{----------------}

% IEEEReview-Jalwana
\cs{Before a detailed discussion of the proposed algorithm, we briefly compare it with the closest available scheme of Kariyappa et. al \cite{kariyappa2019improving}. 
Similar to our work, they explored the role of gradient disparity in promoting adversarial robustness. 
% Our novel formulation improves robustness of a model by the induced orthogonal gradients. 
% Recently, Kariyappa et. al \cite{kariyappa2019improving} has explored the role of gradient disparity in promoting adversarial robustness. 
However, their algorithm modifies the model gradients with strict adherence to linear dependency of the gradients. This contrasts with the central theoretical argument of our method i.e., the linear independence of gradients.  
Moreover, their formulation is particularly tailored to the robustness of model ensembles that requires the specific modification of all the non-linearities that are present in the models. 
% While our algorithm is able to improve the accuracy of an individual model and does not require any architectural changes.
Our novel scheme improves the adversarial robustness of an individual model via novel orthogonal gradient regularization without the need of any architectural changes. The details of our scheme are presented next.   }

%\textcolor{red}{AsimSTART}

\newcommand{\Exp}[1]{\underset{#1}{\mathbb E}}
\begin{algorithm}[h]
 \caption{Orthogonal Model Training}
 \label{algo} 
 \begin{algorithmic}[1]
 \renewcommand{\algorithmicrequire}{\textbf{Input:}}
 \renewcommand{\algorithmicensure}{\textbf{Output:}}
 \REQUIRE  Classifier $\mathcal M$, reference classifier $\mathcal K$, training samples $\overline{\mathcal D}$,
 validation samples $\mathcal D$, 
 mini-batch size $b$, orthogonal parameter $\gamma$, loss-function $\mathcal L$
 \ENSURE $\mathcal M$.
 \STATE Initialize $\boldsymbol{g_1}$, $\boldsymbol{g_2}$  to $\boldsymbol{0} \in \mathbb R^{b \times d}$ and $\wp^*= \delta=0$,  $\mathcal M$ parameters $(\theta^*) $  randomly, epoch = 1, $Acc_t=Acc_{t-1}=0$
\WHILE {$ Acc_{t} >  Acc_{t-1}$} 
    \WHILE {$i,l \sim \overline{\mathcal D}$} 
%     \STATE $i,l \sim \overline{\mathcal D}$, \hspace{2mm} s.t. $|i| = b-1$
    \STATE $\boldsymbol{g}_1 = $ $\frac{\nabla_i\mathcal L(\mathcal K(i),l)}{||\nabla_i\mathcal L(\mathcal K(i),l)||_2}$
    \STATE $\boldsymbol{g}_2 = $ $\frac{\mathcal L(\mathcal M(i),l)}{||\mathcal L(\mathcal M(i),l)||_2}$
    \STATE $\delta = \mathbb E[\boldsymbol{g}_1 \odot \boldsymbol{g}_2]$ 
     \STATE $\wp^*  = \mathbb E[\mathcal L(\mathcal M(i),l)] + \lambda\delta$
    % \STATE $\wp^*  = \frac{1}{\underset{{\boldsymbol{(i,l)}\sim \mathcal{D}}}{\sum} L(\mathcal M(i),l)} \mathcal L(\mathcal M(i),l) + \lambda\delta$
    \STATE $\theta^* \gets$ Gradient descent on $\wp^*$
    \ENDWHILE
      \STATE epoch = epoch + 1
    \IF {epoch$~\%~20 = 0$} 
    	\STATE $Acc_{t-1} = Acc_{t}$
        \STATE $Acc_t \gets $ Find Accuracy on $\mathcal D$
    \ENDIF
 
\ENDWHILE
  \STATE return 
 \end{algorithmic}
 \end{algorithm}

Our procedure to induce misalignment in model gradients is summarized in Algorithm~\ref{algo}. The algorithm solves the optimization problem in Equation~\ref{orthoTrainingEq} with a guided gradient descent strategy. 
Mini-batches of the training samples are employed for a multi-step traversal of the enhanced loss surface. 
 Iterative stepping in the direction of decreasing  `$\wp^*$' with gradient descent allows the model to gain classification accuracy along with orthogonal alignment of its gradients w.r.t.~the reference model. 
 Below we describe the procedure in detail, following the sequence in Algorithm \ref{algo}.
 
 We train an orthogonal model $\mathcal M$ (w.r.t.~reference classifier  $\mathcal{K}$), expecting the inputs mentioned in Algorithm \ref{algo}.
%  The inputs include a reference model $\mathcal K$, dataset ($\overline {\mathcal D}$ and $\mathcal D$) and hyper-parameters that include mini-batch size $b$ and orthogonal parameter $\lambda$. 
 Briefly ignoring the initialization on line~1 and criterion on line~2, the algorithm first samples a mini-batch of size `$b$'.
 On line~4-5, gradients of classification loss function w.r.t samples of mini-batch are calculated for the training and  reference models. 
 The gradients are cast as  $\mathbb R^d$ vectors and normalized by their $l_2$ norms which not only helps in confining their dynamic range to the meaningful interval [0,1], but also allows direct similarity computation under Equation~\ref{dotProductEq}.   
 %The confinement also enables the similarity metric to stay comparable for different mini-batches. 
 Given the normalized gradients, we estimate the similarity `$\delta$' as mean of the gradient dot products as given on line 6, where $\mathbb{E[.]}$ is the Expectation operator.  
We scale `$\delta$'  by a  hyper-parameter `$\lambda$' and update the general classification loss on line 7.  
This hyper-parameter enables explicit control over the degree of gradient misalignment. 
For $\lambda=0$, the algorithm converges to usual training of deep model, ignorant of any gradient alignment. 
For higher values ($\lambda > 30$), the model becomes nearly orthogonal to the  reference model. 

Finally, the optimization algorithm is deployed for updating the weights of model parameters with respect to the gradient of our enhanced loss function. 
The choice of optimization algorithm is not restricted in any manner, however, for a fair evaluation, we employ the same optimization algorithm as used in the training of our reference models.  
The algorithm continues to improve the validation accuracy using the training data.
Due to the stochastic  nature of deep models, we monitor the accuracy of the model after every 20 epochs, as indicated on line 11-13. This is a purely empirical strategy to automate the procedure and can be replaced by manual monitoring. %We choose this setup for automated procedure.       

In general, defenses against adversarial attacks are well known to reduce the performance of the original model \cite{szegedy2013intriguing,goodfellow2014explaining,moosavi2016deepfool,akhtar2018threat, miyato2016adversarial,madry2017towards,woods2019reliable, bai2017alleviating, gu2014towards, gao2017deepcloak, ross2018improving,lyu2015unified,shaham2018understanding,woods2019reliable}.
Hence, before we provide the actual quantitative results in Section \ref{sec:Exp}, we find it necessary to discuss the intuition behind the attractiveness of our proposal of employing model orthogonality as a defense. 
Below, we give a simplified explanation that reveals how two  models can still have very similar classification performance, despite their gradients beings orthogonal.

Deep classification models can be viewed as stack of non-linear transformation layers where each internal layer sequentially transforms its input before feeding it to the next layer. 
Consequently, the model as a whole is able to project an input to the output - class label in the case of classifiers.
For an input image $i \in \mathbb R^d$, the output of a deep model, say $i'$ can be expressed as:
\begin{align}
    i' = T_N(T_{N-1}(T_{N-2}(\dots~T_2(T_1(i))))),  
\end{align}
where `$N$' is the number of layers.
Our argument is that despite large differences between the intermediate projections by the internal layers of two misaligned models, both models  can still achieve similar classification objective.   
Thereby, enabling both models to preserve the same desired classification accuracy.  

\begin{figure}[t]
\centering
\includegraphics[width=0.35\textwidth]{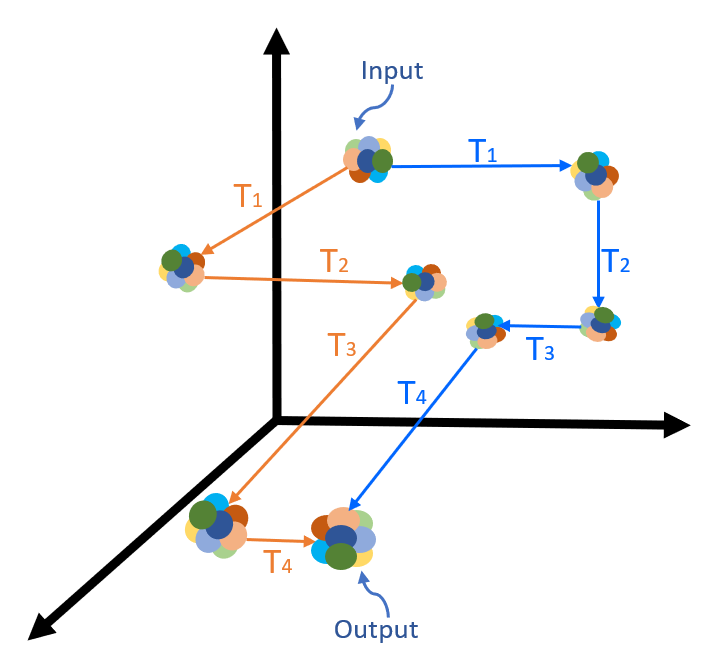}
\caption{Two set of dissimilar transformations shown in blue and orange, are applied to a set of points. Each transformation is represented by $T_x$. Despite being pair-wise orthogonal the transformations eventually  project the input points to the same output.}
\label{highDimensionalTrans}
\end{figure}

We provide an intuitive explanation of this insight by considering two simple linear models composed of four transformation layers ($T_1 \dots T_4$). 
For the ease of understanding, we restrict the input and transformation spaces to $3$ dimensions. 
In Fig.~\ref{highDimensionalTrans}, we illustrate the transformations by the internal layers of the models by arrows of different colors.
It can be observed that despite orthogonality between each pair of corresponding transformations, the input points are eventually mapped to the same output by both sets of transformations (i.e.~models).
In theory, this identifies the possibility of achieving the exact same performance by two models despite orthogonality between their internal layers.

In the above discussion we consider the simple case of 3 dimensional space with linear systems. For our argument, this is actually a more constrained setup as compared to higher dimensional spaces that also allow non-linear systems. In that setting, the flexibility of non-linearity and inherent low correlation between random vectors provide a more conducive setup for our argument to hold. 
In high dimensional vector spaces, a slight variation in the components of a vector can lead to a drastic change in its orientation. 
In fact, in such spaces the probability of random vectors to be correlated approaches to zero as the dimensions become very large~\cite{gorban2018blessing}. We refer interested readers to Gorban et. al \cite{gorban2018blessing} for more details regarding this phenomenon. 
Here, we capitalize on this phenomenon - which is commonly seen as a curse of dimensionality - constructively.  

%We further support this point with empirical evidence in Section~\ref{sec:Exp}.

%Each model, shown in a different color, is able to project in entirely different manner before yielding a similar final output as illustrated in Figure \ref{highDimensionalTrans}. 
%This ease of models to achieve similar classification results despite disparate projections paths hints towards nominal decrease in standard classification accuracy for the orthogonal models. 
%Detailed empirical evidence is presented in next section.

% We argue that gradient misaligned variants of a model may also achieve similar classification accuracy beside different projection behaviour of individual layers.   

% Deep classifiers comprise `$N$' non-linear layers, say $T_n$, that  hierarchically transform the inputs to the final prediction scores. 
% For any input $i \in \mathbb R^d$, the output of a deep model $i'$ can be expressed as
% \begin{align}
%     i' = T_N(T_{N-1}(T_{N-2}(\dots~T_2(T_1(i))))). 
% \end{align}For simplicity and ease of visualization, two models with four linear-transformations considered in 3D subspace are shown in Figure \ref{highDimensionalTrans}. 

\begin{figure}[h]
\centering

\includegraphics[width=0.45\textwidth]{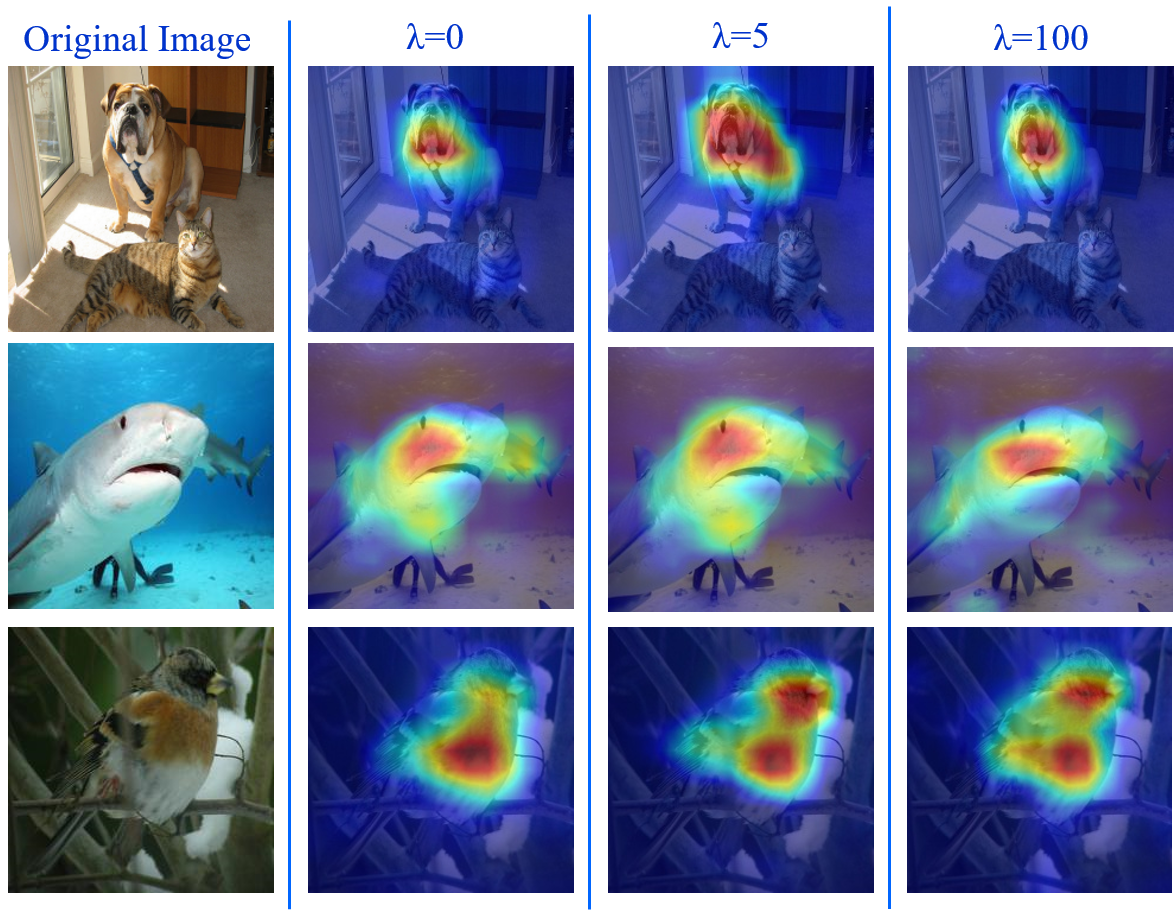}
\caption{Localization of salient regions of input images by Grad-CAM \cite{selvaraju2017grad}. $\lambda = 0$ indicates the reference VGG-16 model and larger `$\lambda$' values identify models with proportionally dissimilar gradients w.r.t.~the reference VGG-16. For the randomly chosen images, all models result in correct predictions and also focus on similar regions to make their decision despite large dissimilarity between their gradients.}
\label{gradCAMResults}

\end{figure}

\begin{figure*}[h!]
\centering
\includegraphics[width=\textwidth]{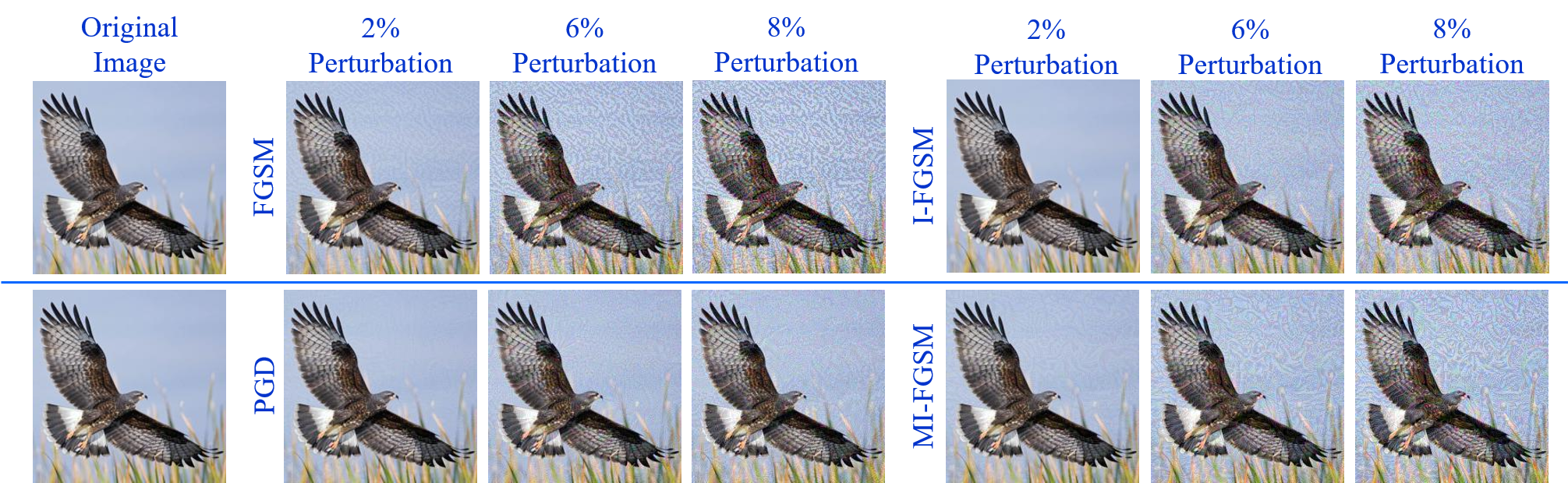}
\caption{Perceptibility of adversarial perturbations is shown for different attack schemes as we vary the hyper-parameter `$\epsilon$' in attack schemes. Higher value of `$\epsilon$' results in more visual distortions. }
\label{perturbationPerceptibility}
\end{figure*}

Expanding on the implications of our argument, misaligned gradients naturally raise  concerns about differences in the internal reasoning of the  models. 
% \textbf{Variation in Explanatory regions of Models: }
% Deep models are often considered black-boxes due to 
%  sheer volume of parameters and a variety of non-linear layers. 
With sheer volume of parameters and variety of non-linear layers, the direct analysis of causality between input and output of deep models is not possible. 
% We resort to explanatory technique, GradCAM, that enables reliable qualitative insights to the inner functionality of models \cite{adebayo2018sanity}. 
% Among these, schemes that identify regions that play salient role in classification decisions are considered reliable \cite{adebayo2018sanity}. 
Hence, we analyze this reasoning for ordinary and orthogonal models via a popular explanatory algorithm called GradCAM~\cite{selvaraju2017grad}. 
This technique exploits the internal representation of a model to reliably localize the region of interest in the input image for the model~\cite{adebayo2018sanity}. 
In the presence of dissimilar  gradients, one may anticipate a significant difference between these regions for ordinary and orthogonal models, given the same input.  

We show representative results of Grad-CAM for different VGG-16 models  with varying degree of orthogonality in Fig.~\ref{gradCAMResults}. 
The selected images, shown in the first column, are from random classes, and are  predicted correctly by the used models.
The images with heat maps from Grad-CAM, where the hot spots identify the region of interest for a model. Here, $\lambda = 0$ indicates the reference VGG-16 model and larger `$\lambda$' values identify models with proportionally large  dissimilarity of gradients  with the reference model, as explained above.
We provide complete details of our experiments in the next Section. Here, we highlight with Fig.~\ref{gradCAMResults} that despite large dissimilarity between the gradients of models, they are generally able to identify the same region of interest in the images. 
%salient regions qualitatively stay similar. 

For the top row with dog image, all three models consider dog's  face as the most salient region. Similarly, eye and nose are considered important by the models for the middle row with  shark image. 
For the bird image, shown in the last row, despite significant visual differences in the GRAD-CAM outputs, it is clearly identifiable that the head and some parts of the body are used by all the models for correct classification. These images ascertain that despite large dissimilarities between the gradients of different models (with the same architecture), they are still able to perform similar internal reasoning regarding the semantics of the inputs to perform correct classification.
Hence, besides the potential to achieve the desired high accuracy of the reference model, orthogonal models are also able to preserve the intuitive internal reasoning of the models.
The proposed models are able to significantly reduce the transferability of adversarial attacks while maintaining these desirable properties.  
%The intuitive reasons, higher dimension probability and explanatory algorithms allude towards the ease of gradient misalignment in deep models.  
In the next section, we focus on providing thorough empirical evidence that gradient misalignment can boost immunity against black-box adversarial attacks. 

% Surprisingly, as Figure \ref{gradCAMResults} demonstrates, we observe that the prominent region's position and silhouette stays similar.

% Similarity in regions signifies the reasoning resemblance in orthogonal model and ordinary model. 
% For example the top row shows that ordinary classifier considers the facial region of the dog as the most important part to classify the breed.
% However, even with nearly independent alignment of gradients, this qualitative aspect of orthogonal model reasoning stays the same.  
% It is argued that this similarity is inline with the illustrated projection view of deep models in Figure \ref{highDimensionalTrans} and therefore an aftermath of hyper dimension of gradients subspace and .

% Ease of achieving orthogonality plays an important role to pres
% role to preserve the localization of significant regions, which in turn upholds the classification accuracy. As a positive side-effect the model gains immunity against the adversarial attacks. 

%-------------------------------------------------------------------------

\begin{figure*}[t]
    \centering
    \includegraphics[width=\textwidth]{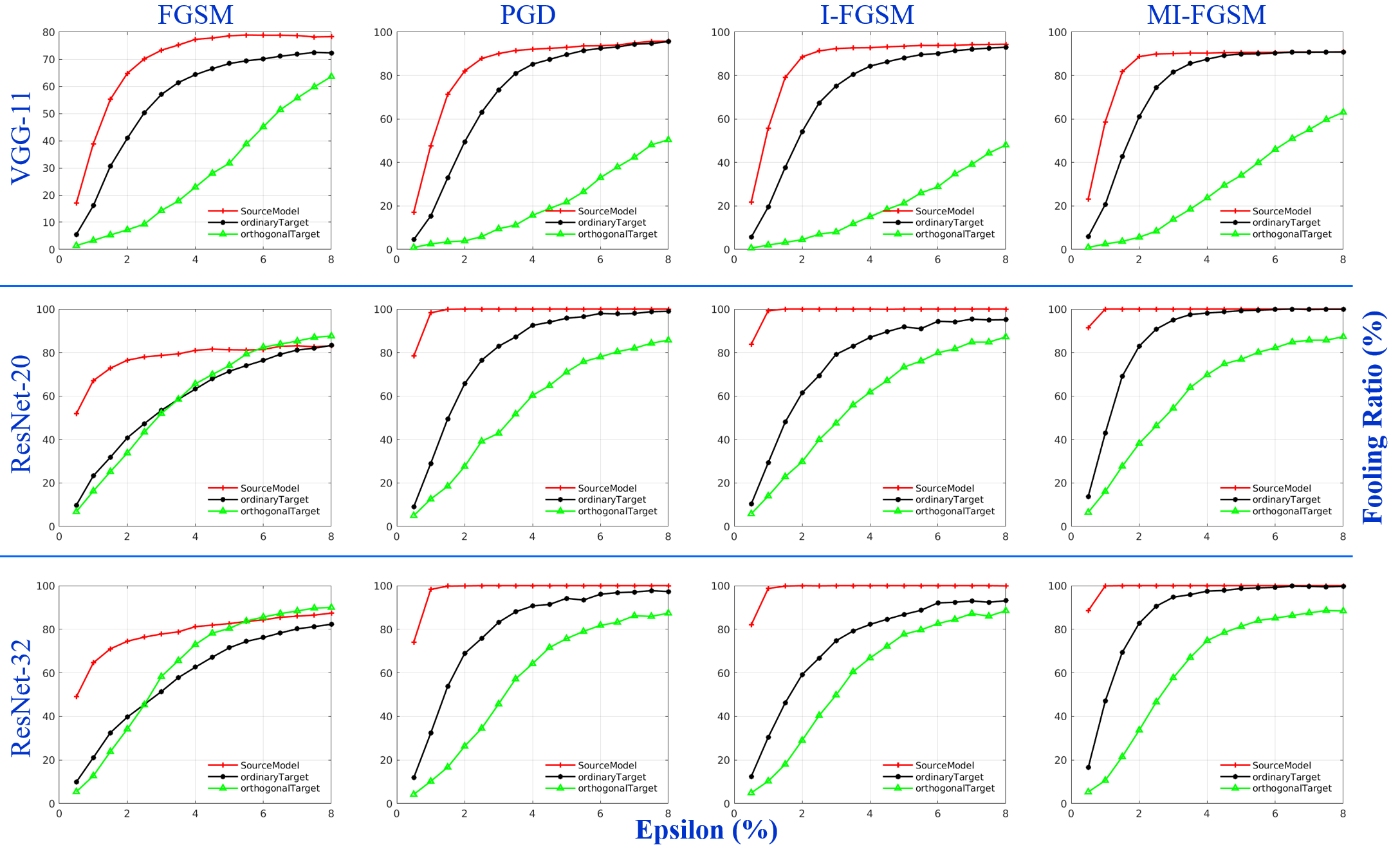}
    \caption{Fooling ratio ($\%$) shown on the vertical axis for the perturbations crafted on ordinary classifier (source in \textcolor{red}{red}) and transferred to a retrained ordinary target model (black) and orthogonal target model (\textcolor{green}{green}). The model shown are trained on CIFAR-10 dataset. The horizontal axes show the step size ($\epsilon$) for perturbation computation varied from 0 to 8$\%$ of the image dynamic range.}
    \label{robustnessFigCIFAR10}
\end{figure*}

\section{Experiments }
\label{sec:Exp}
We first outline the experimental setup before detailed discussion and analysis of  empirical evaluation of our technique on different datasets. 
% We evaluate the robustness of orthogonal models, in comparison to ordinary, against black-box adversarial attacks. Our study includes several architectures trained over CIFAR-10 and ILSVRC ImageNet 2012 dataset. The experimental setup is based on the Pytorch framework \cite{NEURIPS2019_9015} as its native support for dynamic graphs provides convenience for gradient regularization. 

We evaluate the robustness of gradient misaligned models against black-box adversarial attacks. 
These attacks assume that the internal gradients of target model are not available. Therefore,  perturbations are crafted using  surrogate models and transferred to the target models. 
Observing success of transferred perturbations (fooling ratio) provides a fair estimate of model immunity.  
We demonstrate the generalization of our approach by including several architectures trained over CIFAR-10 \cite{krizhevsky2009learning} and ILSVRC ImageNet 2012 dataset~\cite{deng2009imagenet}. 

% We evaluate our technique for defense against black box adversarial attacks where no internal information of target model is available.
% Following the standard norm \cite{liu2016delving}, we craft perturbation over an independently trained model and transfer it to our models. 
% Observing fooling ratio of transferred perturbations provides a fair estimate of model immunity.  
% To demonstrate the generalization of our approach we include  several architectures trained on CIFAR-10 and ILSVRC ImageNet 2012 dataset. 
% Our experiments setup is based on the Pytorch framework \cite{NEURIPS2019_9015}  as its native dynamic graphs provides convenience for gradient regularization. 

We consider two separate choices of source models to craft perturbations. 
First we keep similar source and target model architectures.
This enables us to verify robustness under the more challenging condition as it is known that perturbations transfer maximally to similar architectures \cite{liu2016delving}. Later, we consider the case of varying source architecture to validate robustness against diverse perturbations. 
To evaluate our models, for each source-target model pair, we sample $1000$ images from the  validation sets that are correctly classified i.e.~prior baseline accuracy for models is 100$\%$. 
These images are subsequently perturbed and classified by the  target model.  
Hence, the bias in fooling ratio due to already misclassified samples is completely  avoided in our results.

We craft adversarial perturbations using four well-known attack schemes, namely
the single-step FGSM~\cite{goodfellow2014explaining} and multi-step PGD~\cite{madry2017towards}, I-FGSM~\cite{kurakin2016adversarial} and MI-FGSM~\cite{dong2018boosting}. 
These techniques generate strong transferable perturbations and are often used to validate the robustness of deep models\cite{shi2019curls, madry2017towards}.
We used their publicly available implementations in  Foolbox\cite{rauber2017foolbox} and AdverTorch \cite{ding2019advertorch}. 
Our experiments are based on the Pytorch framework~\cite{NEURIPS2019_9015}  as its native dynamic graphs provide  convenience for our gradient regularization.

The most salient qualitative feature of a perturbation is its visual footprint.
Perturbed images that are easy to identify by human visual inspection are of little practical concern.  
Therefore, following the standard practice, we keep the perceptibility of perturbations in an acceptable range by controlling the step size ($\epsilon$) in attack schemes.   
In Figure~\ref{perturbationPerceptibility}, we illustrate the visual quality of distortions as $\epsilon$ is varied.
It can be observed that perturbations become  more visible with increasing $\epsilon$. 
In our evaluation, we vary this parameter from $0.5\%$ to $8\%$ to demonstrate the efficacy of our scheme. Our coverage of robustness evaluation is more comprehensive because, given the visual perceptibility of distortions, the existing literature generally uses $4$-$5\%$ as the upper bound for perturbations~\cite{moosavi2017universal}, \cite{akhtar2019label}.

\subsection{CIFAR-10 Models}
We first demonstrate the effectiveness of our scheme for the visual models trained over CIFAR-10 dataset \cite{krizhevsky2009learning} which consists of  60,000 $32\times32$ color images uniformly divided into $10$ classes. 

Low computational cost for training models on this dataset allows us to validate our scheme on three different architectures. This includes publicly available VGG-11~\cite{chengyangfu}, ResNet-20~\cite{akamaster} and ResNet-32~\cite{akamaster}. 
The deliberate inclusion of two fundamentally different architectures (VGG and ResNet) permits  us to analyze the  architectural role in developing immunity against adversarial perturbations.
Similarly, inclusion of two variants of ResNet enable us to analyze the role of depth for similar architecture. 
ResNets have a special place in the adversarial arena as recent works have demonstrated that skip connections facilitate the transfer of adversarial examples \cite{wu2020skip}. Hence, we include ResNets in our evaluation.  

% also the  enables us to analyze the role of model's depth in the induced immunity.  
To train our model, we follow the standard practice of reserving 50,000 training samples and 10,000 test samples. 
Multiple randomly initialized ordinary models ($\lambda=0$) and orthogonal models  ($\lambda=30$) were trained via our Algorithm~\ref{algo}. 
In Table \ref{simTableCIFAR10}, we report the notable statistics of these models. 
We provide pairs of models such that `Model-1' is the original model and `Model-2' is its newly trained version. The orthogonal version is distinguished with a superscript `o'.
In the table, it can be observed that when orthogonality is not considered, two  independently trained models have high correlation in their gradients.
Our orthogonality constraint drastically reduces this correlation.
It is also worth noting that the classification accuracy of the models with orthogonality constraint does not reduce much and stays within $5\%$ of the reference model accuracy. This is a considerable improvement over the commonly adopted technique of adversarial training~\cite{athalye2018robustness} that is known to result in 10-20$\%$ reduction in the original accuracy.

% We first demonstrate the effectiveness of our scheme over three different deep visual models trained over CIFAR-10. 
% The low computational cost of dataset allows us to 
% exhaustively test our claims and verify the generalization of our scheme across different models. 
% However, we have not ignored the importance of more challenging and realistic test-bed of ImageNet ILSVRC 2012. 
% Detailed analysis of our framework based on VGG-16 architecture has been included to strengthen our claims. 

% All of the experiments are based on the Pytorch framework \cite{NEURIPS2019_9015}. 
% The dynamic graphs of the framework enables flexible regularization of gradients. 

\begin{table}[t]
\footnotesize
\centering
\begin{tabular}{p{38pt}p{42pt}p{59pt}p{52pt}}
\hline
\textbf{Model-1}         &       \textbf{Model-2}          &  \textbf{Similarity ($\delta$)}     & \textbf{Val. Acc.}\\
\hline
VGG-$11$          &       VGG-$11$           		&  0.3834 $\pm$ 0.03          & 90.5  $\pm$ 0.03 $\%$ \\ 
VGG-$11$          &       VGG-$11^\text{o}$         &  0.0070 $\pm$ 0.0004        & 86.2  $\pm$ 0.06 $\%$ \\ 
ResNet-$20$       &       ResNet-$20$        	    &  0.1338 $\pm$ 0.04          & 91.4  $\pm$ 0.03 $\%$ \\ 
ResNet-$20$       &       ResNet-$20^\text{o}$      &  0.0018 $\pm$ 0.0003        & 88.0 $\pm$ 0.04 $\%$ \\ 
ResNet-$32$       &       ResNet-$32$               &  0.1507 $\pm$ 0.04          & 92.5  $\pm$ 0.02 $\%$ \\  
ResNet-$32$       &       ResNet-$32^\text{o}$      & 0.0031 $\pm$ 0.0003         & 87.0 $\pm$ 0.03 $\%$  \\  

\hline
\end{tabular}
\caption{Statistics of independently trained model pairs on CIFAR-10. For each model training, the weights were  initialized randomly with Xavier initialization \cite{glorot2010understanding}.
Model-1 indicates the reference model. Model-2 is the newly trained model with the same architecture.  The superscript `o' indicates orthogonal model trained with  hyper-parameter $\lambda = 30$. Orthogonality drastically reduces  gradient similarity with the reference model while maintaining the accuracy  within an acceptable range.} 
\label{simTableCIFAR10}
\end{table}

\begin{figure*}[t]
    \centering
    \includegraphics[width=\textwidth]{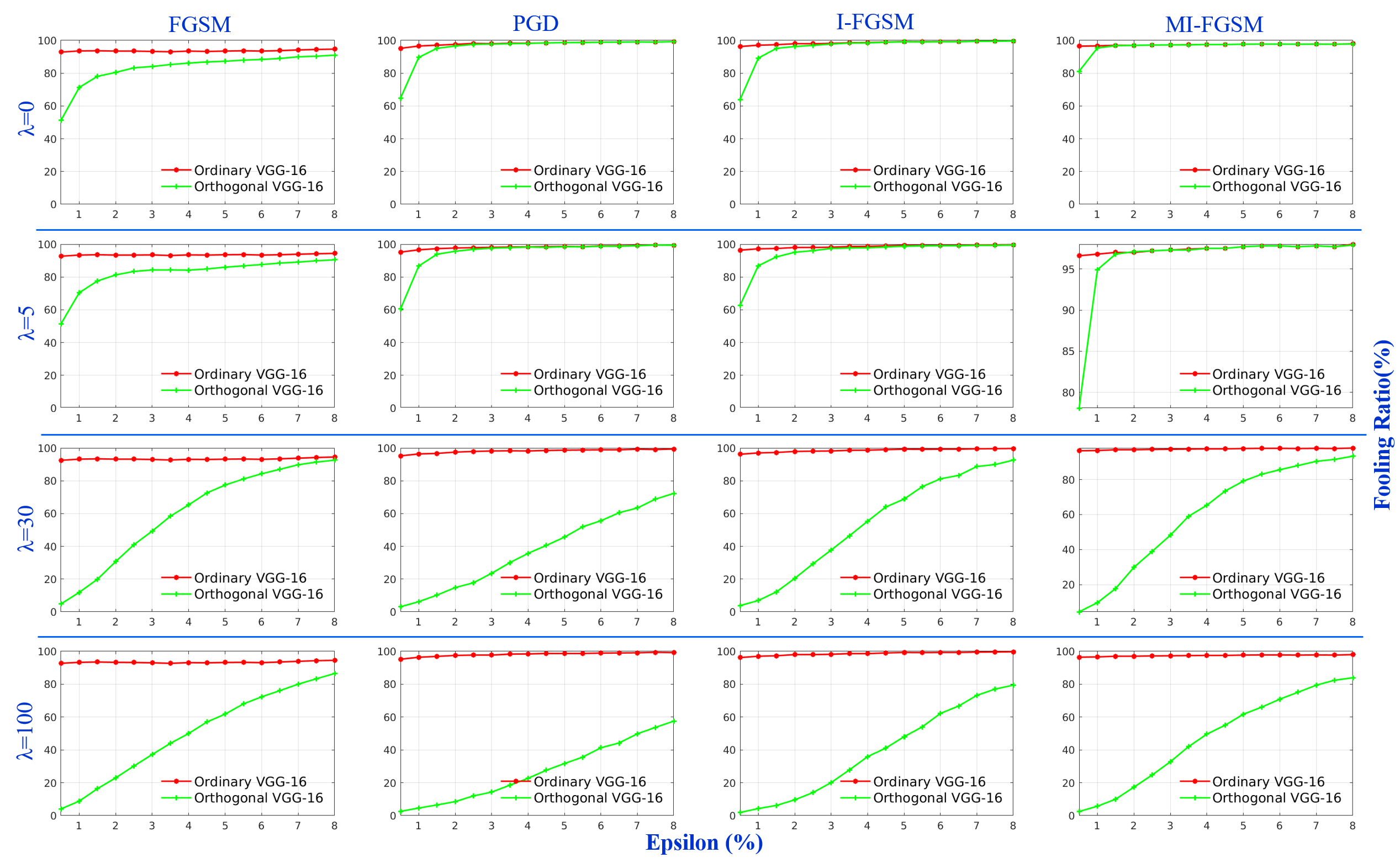}
    \caption{\textcolor{black}{Fooling ratio ($\%$) shown on the vertical axis for the perturbations crafted on ordinary classifier (source in \textcolor{red}{red}) and transferred to orthogonal model (\textcolor{green}{green}). The models shown are trained on ImageNet dataset. The horizontal axis show the step size ($\epsilon$), and is varied from 0 to 8$\%$ of images dynamic range.}}
    \label{robustnessFigImageNet}
\end{figure*}

In Figure \ref{robustnessFigCIFAR10}, we demonstrate the impact of gradient misalignment over  adversarial attack  transferability. 
In the first row of the figure, we use VGG-11 architecture and train a `source model' that is used to craft adversarial perturbations. These perturbations are used to attack three models with the same VGG-11 architecture. The first is the source model itself (red). The second is the `ordinary target' model (black), induced with different initialization of the training process. This setup emulates the commonly encountered black-box attack if the attacker is able to  correctly guess the  (standard) architecture of the target model.
The third model is the `orthogonal target' model (green) that has the VGG-11 architecture, but it is trained with our technique. 
The second and the third rows of the figure show the same setup for ResNet-20 and ResNet-32 architectures. Each column of the figure shows results with different attacks, mentioned at the top of the figure. 
In each plot, vertical axis reports the  fooling rate, while the horizontal axis represents the step size `$\epsilon$' as the  percentage variation of the image dynamic range to compute a perturbation.

%Each row of the figure  includes the results of similar surrogate and target architecture as indicated on the left. 
%Perturbations are crafted on source model (red) and then evaluated on retrained ordinary model (black) and orthogonal (green) variant. 
%In each graph, vertical axis reports the  fooling rate, while the horizontal axis represents the step size $\epsilon$.
% We vary the $\epsilon$ between $0.5\%$ and $8\%$ with a step size of 0.5, therefore sixteen experiments are performed for each type of attack and model type. 

The plots in Figure  \ref{robustnessFigCIFAR10} %Ajmal: check figure
clearly indicate that our orthogonal models have generally stronger resistance against attacks compared to the source and its retrained variants. 
The robustness is significantly higher for VGG architecture than for the ResNets.
We conjecture that this phenomenon results from skip-connections that are known to facilitate the tranferability of perturbations. Nevertheless, the orthogonal ResNets do  exhibit significant immunity against PGD, I-FGSM and MI-FGSM attacks. 
This demonstrates that our technique is able to bring robustness to ResNets as well, despite their inherent facilitation to black-box attacks. 
\cs{Among all the attacks, MI-FGSM is known to generate highly transferable perturbations \cite{dong2018boosting}. However, our orthogonal training enhances the robustness by $55.8\%$ for VGG-11, $22.4\%$ for ResNet-20 and $17.4\%$ for the ResNet-32 against the attacks generated with $\epsilon=0.05$. These results are relative to the ordinary retrained model.}

\subsection{ImageNet ILSVRC2012}

We extend the validation of our approach to the large-scale ImageNet ILSVRC 2012 dataset~\cite{deng2009imagenet} which consists of 1.2 million color images with $224\times224$ resolution and 1000 classes of daily-life objects. 
It is a common practice in adversarial defense literature to perform experiments only with CIFAR-10 and small datasets, e.g.~MNIST~\cite{lecun-mnisthandwrittendigit-2010}. However, for a more comprehensive evaluation, we  present results for this large-scale dataset despite heavy  computational requirements. We prefer this because results on ImageNet are more likely to generalize well to other models in the era of large-scale datasets.  

Recent works \cite{simon2018adversarial} have established a one-to-one relationship between the dimensions of model inputs and adversarial susceptibility of the  models. This makes ImageNet classifiers   especially challenging to defend against   adversarial perturbations as compared to MNIST and even CIFAR models.  While the lower computational complexity of CIFAR-10 dataset permits us to train numerous models from scratch, ImageNet training of multiple models from scratch is computationally prohibitive.

% ~~~~The higher resolution of input images and large number of classes makes ImageNet ILSVRC2012 a far more challenging test-bed to study the adversarial perturbations \cite{simon2018adversarial}. 
% Lower computational complexity of CIFAR-10 dataset permits to train the models from scratch in matter of hours. 
% However for ImageNet, training of models from scratch is computationally prohibitive. 

\begin{table}[h]
\footnotesize
\centering
\begin{tabular}{p{0.8cm}p{2cm}p{2cm}p{2cm}}
\hline
\textbf{Lambda} & \textbf{Similarity} & \textbf{Train. Time}  & \textbf{Validation} \\  
$\lambda$ 	   & $\delta$ 				         & 		\textbf{(Per Model)} 			      & 		\textbf{Accuracy} 		\\ \hline
	0   	   & 	0.522 $\pm$ 0.0516     & 	   	   190 $\pm$ 10 mins     	&   		58.05$\pm$ 2$\%$ 		\\
	5   	  &    0.351 $\pm$ 0.0203     & 			   70 $\pm$ 3 hrs         		&   	   57.05$\pm$ 2$\%$  		\\
	30  	 & 	  0.012 $\pm$ 0.0091     & 			  70 $\pm$ 3 hrs        		   &   		  55.01$\pm$ 2$\%$ 		\\
	100 	&    0.002 $\pm$ 0.0008     & 			 70 $\pm$ 3 hrs         		  &   		 53.17$\pm$ 2$\%$ 			\\
\hline
\end{tabular}
\caption{Statistics of the trained models with varying degree of similarity with baseline. The similarity is explicitly controlled with the hyper-parameter $\lambda$. First row  captures results for different initializations of the baseline. For $\lambda > 0$, similarity is always evaluated with respect to the baseline model. Results are for ImageNet validation set using VGG-16 architecture.} 
\label{simTable}
\end{table}

\begin{figure*}[t]
    \centering
    \includegraphics[width=\textwidth]{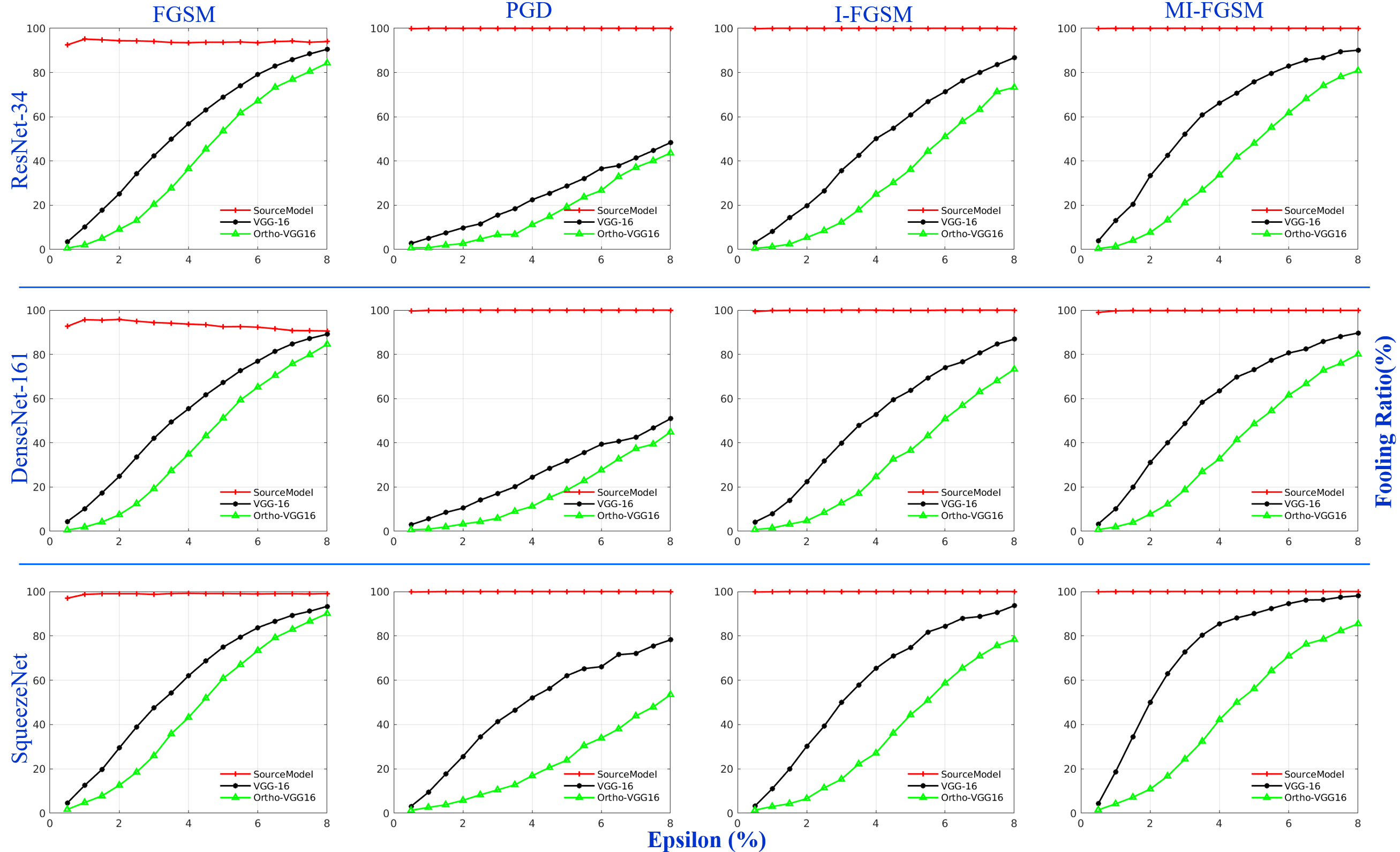}
    \caption{Fooling ratio  ($\%$) is shown on the vertical axis for the perturbations crafted on ordinary classifier (source in \textcolor{red}{red}) and transferred to ordinary VGG-16 model (black) and orthogonal VGG-16 model (\textcolor{green}{green}). The horizontal axis show the step size ($\epsilon$), and is varied from 0 to 8$\%$ of images dynamic range. Each row reports the results of perturbations crafted on a pretrained ImageNet model as shown in blue font on the left hand side. }
    \label{transferabilityFromOtherModelsFig}
\end{figure*}

For computational reasons, we restrict our evaluation for ImageNet to VGG-16 architecture and employ the validation set of ImageNet for experiments. This set comprises 50,000 images. We follow the standard split of 80/20 for the training and testing our models. 
Ordinary models, i.e.~base-line, are prepared by fine tuning the ImageNet pretrained models available in Pytorch \cite{NEURIPS2019_9015}. Fine tuning is performed by following the standard procedure of re-initialization and training of the last fully connected layers. The initial layers are known to capture the common shared concepts and relearning the last few layers suffices the model to capture the underlying high level features of the new distribution.

We use the baseline model as the reference model and train numerous other models by varying the hyper-parameter `$\lambda$' in Algorithm~\ref{algo}. 
% We train a number of models by varying the hyper parameter $\lambda$ in Algorithm \ref{algo} and using baseline model as reference model.
Each experiment is  repeated five times for which the important statistics are summarized in Table~\ref{simTable}. Here, the similarity of models for each `$\lambda$' is w.r.t.~the baseline model. It can be observed that in simple retraining i.e. $\lambda=0$, strong correlation exists among the gradients. As `$\lambda$' increases to $100$, the correlation almost vanishes. It is worth noting that similar to CIFAR-10 models, the standard classification accuracy suffers slightly and at the near orthogonal alignment it stays within $5\%$ of the original model's accuracy. 
We have included the training time of the models to show the computational complexity. Generally it takes $3$ days to train an orthogonal model on ImageNet validation set using NVIDIA Titan V GPU with 12GB RAM. 

In Figure \ref{robustnessFigImageNet}, we illustrate the role of dissimilarity in boosting model robustness against transferable attacks. 
Perturbations are crafted on source model, shown in red, and then evaluated on orthogonal model, represented in green. 
Each column presents the results for a particular attack type and each row presents the results for different correlations between the models as controlled by `$\lambda$' value.
The vertical axes reports the fooling rate of the classifiers and horizontal axes represents the step size `$\epsilon$' as percentage of the original dynamic image range. 
It can be observed that for each attack type (column), as we increase the misalignment of gradients (from top to bottom row) the robustness of orthogonal model increases. 
Overall, orthogonal models show robustness to all types of attacks. However, maximal resistance is observed for PGD which is considered to be the strongest iterative attack \cite{madry2017towards}. \cs{The orthogonal model increases the robustness by $67\%$ for the PGD attacks generated with $\epsilon=0.05$.}

\subsection{Attacks from Other Models}

The above results clearly establish the efficacy of our scheme against perturbations generated from similar architectures. 
% "We further demonstrate that the proposed constraint makes the resulting orthogonal model even more robust to perturbations that are crafted from dissimilar architectures, a more realistic setting for black-box attacks."
{We further demonstrate that our scheme enhances the resulting orthogonal model robustness to perturbations crafted from dissimilar architectures - a more realistic setting for black-box attacks.}
% We further demonstrate that the gained immunity becomes more of a generic trait of orthogonal models by observing its response against perturbations crafted from dissimilar architectures.%Ajmal: you are not claiming much credit in the last sentence

In Fig.~\ref{transferabilityFromOtherModelsFig}, we show how attacks computed from different source model transfer to our orthogonal VGG-16 model.
Each row of the figure shows the fooling results of our orthogonal target for perturbations crafted by the architecture mentioned on the left.  Each column shows the results for a single attack scheme as indicated on the top. 
The plot axes follow our conventions from the previous figure. 
The red curve represents the fooling ratio on source model and the green curve represents the same for orthogonal VGG-16 model. 
For clear benchmarking, fooling ratio for the ordinary VGG-16 model (not made orthogonal) is included as black curves.

As can be seen, we use three diverse architectures to craft the perturbations, including  ResNet-34~\cite{he2016deep}, DenseNet-161~\cite{huang2017densely} and SqueezeNet~\cite{iandola2016squeezenet}. We used the  pretrained models of these architectures available in Pytorch~\cite{NEURIPS2019_9015}. 
In Fig.~\ref{transferabilityFromOtherModelsFig}, the graphs illustrate superior robustness of our orthogonal model in comparison to the ordinary one. 
It can be observed that the green curve reports lower fooling ratio by staying reasonably below the black curve for all combinations of attack schemes and source models. 
% It can be observed that for all combinations of attack scheme and source model, the green curve reports lower fooling ratio by staying reasonably below the black curve. 
These results support that controlled misalignment of model gradients can boost  immunity of models against  adversarial perturbations especially in a black-box setting when the target model architecture is unknown.

\begin{table*}[t!]
\resizebox{\textwidth}{!}{%
\begin{tabular}{llc|c|c|c|c|}
\cline{4-7}
\multicolumn{1}{r}{} &  & \multicolumn{1}{l|}{} & \multicolumn{4}{c|}{\cellcolor[HTML]{EFEFEF}\textbf{Attacks}} \\ \hline
\rowcolor[HTML]{EFEFEF} 
\multicolumn{1}{|l|}{\cellcolor[HTML]{EFEFEF}\textbf{Models}} & \multicolumn{1}{l|}{\cellcolor[HTML]{EFEFEF}\textbf{Defense}} & \multicolumn{1}{l|}{\cellcolor[HTML]{EFEFEF}\textbf{\begin{tabular}[c]{@{}l@{}}Clean Samples\\ Fooling ratio\end{tabular}}} & \multicolumn{1}{c|}{\cellcolor[HTML]{EFEFEF}\textbf{\begin{tabular}[c]{@{}c@{}}FGSM\\ Fooling ratio\end{tabular}}} & \multicolumn{1}{c|}{\cellcolor[HTML]{EFEFEF}\textbf{\begin{tabular}[c]{@{}c@{}}PGD\\ Fooling ratio\end{tabular}}} & \multicolumn{1}{c|}{\cellcolor[HTML]{EFEFEF}\textbf{\begin{tabular}[c]{@{}c@{}}I-FGSM\\ Fooling ratio\end{tabular}}} & \multicolumn{1}{c|}{\cellcolor[HTML]{EFEFEF}\textbf{\begin{tabular}[c]{@{}c@{}}MI-FGSM\\ Fooling ratio\end{tabular}}} \\ \hline
\multicolumn{1}{|l|}{} & \multicolumn{1}{l|}{No defense} & 0 & 73.3 / 78.6 & 89.8 / 93.2 & 92.4 / 93.5 & 90.1 / 90.6 \\ \cline{2-7} 
\multicolumn{1}{|l|}{} & \multicolumn{1}{l|}{JPEG-Quality=95} & 0.8 & 70.2 / 77.1 & 88.4 / 93.0 & 91.3 / 93.5 & 90.0 / 90.5 \\ \cline{2-7} 
\multicolumn{1}{|l|}{} & \multicolumn{1}{l|}{JPEG-Quality=90} & 1.1 & 67.8 / 76.3 & 86.2 / 92.9 & 89.8 / 93.3 & 89.7 / 90.5 \\ \cline{2-7} 
\multicolumn{1}{|l|}{} & \multicolumn{1}{l|}{JPEG-Quality=85} & 1.6 & 65.3 / 75.0 & 83.5 / 92.5 & 86.5 / 92.6 & 87.5 / 90.5 \\ \cline{2-7} 
\multicolumn{1}{|l|}{} & \multicolumn{1}{l|}{JPEG-Quality=80} & 1.9 & 62.0 / 73.4 & 79.3 / 91.6 & 81.5 / 91.8 & 84.9 / 90.5 \\ \cline{2-7} 
\multicolumn{1}{|l|}{} & \multicolumn{1}{l|}{TVM-Weight=3} & 1.3 & 49.0 / 66.8 & 51.4 / 81.9 & 51.7 / 80.8 & 87.5 / 90.4 \\ \cline{2-7} 
\multicolumn{1}{|l|}{} & \multicolumn{1}{l|}{TVM-Weight=5} & 2.7 & 65.4 / 77.3 & 74.9 / 90.2 & 75.9 / 89.9 & 82.1 / 90.4 \\ \cline{2-7} 
\multicolumn{1}{|l|}{} & \multicolumn{1}{l|}{TVM-Weight=7} & 3.9 & 60.6 / 74.7 & 66.8 / 87.0 & 65.4 / 83.4 & 76.2 / 89.0 \\ \cline{2-7} 
\multicolumn{1}{|l|}{} & \multicolumn{1}{l|}{TVM-Weight=9} & 5.6 & 56.5 / 72.2 & 59.2 / 82.0 & 56.9 / 77.0 & 68.6 / 86.3 \\ \cline{2-7} 
\multicolumn{1}{|l|}{} & \multicolumn{1}{l|}{Bit Reduction - Depth=1} & 43.9 & 62.3 / 68.2 & 61.2 / 69.9 & 62.0 / 69.6 & 64.1 / 73.0 \\ \cline{2-7} 
\multicolumn{1}{|l|}{} & \multicolumn{1}{l|}{Bit Reduction - Depth=2} & 15.0 & 61.5 / 69.9 & 69.3 / 87.4 & 73.5 / 87.3 & 77.9 / 89.2 \\ \cline{2-7} 
\multicolumn{1}{|l|}{} & \multicolumn{1}{l|}{Bit Reduction - Depth=3} & 1.6 & 69.0 / 76.6 & 86.5 / 92.9 & 90.1 / 93.2 & 89.3 / 90.5 \\ \cline{2-7} 
\multicolumn{1}{|l|}{} & \multicolumn{1}{l|}{Bit Reduction - Depth=4} & 0.3 & 72.8 / 77.9 & 89.3 / 93.1 & 92.3 / 93.5 & 90.1 / 90.6 \\ \cline{2-7} 
\multicolumn{1}{|l|}{} & \multicolumn{1}{l|}{Bilateral Filtering - Window=3} & 14.5 & 51.7 / 65.5 & 57.1 / 82.0 & 55.5 / 75.0 & 64.9 / 85.8 \\ \cline{2-7} 
\multicolumn{1}{|l|}{} & \multicolumn{1}{l|}{Bilateral Filtering - Window=5} & 17.2 & 48.8 / 62.1 & 47.6 / 67.8 & 45.6 / 60.4 & 53.1 / 78.2 \\ \cline{2-7} 
\multicolumn{1}{|l|}{} & \multicolumn{1}{l|}{Bilateral Filtering - Window=7} & 13.2 & 47.9 / 61.9 & 45.3 / 66.8 & 43.6 / 58.7 & 52.5 / 77.1 \\ \cline{2-7} 
\multicolumn{1}{|l|}{} & \multicolumn{1}{l|}{Bilateral Filtering - Window=9} & 13.6 & 46.7 / 60.8 & 44.3 / 63.7 & 41.7 / 57.0 & 49.6 / 74.8 \\ \cline{2-7} 
\multicolumn{1}{|l|}{\multirow{-18}{*}{\textbf{\begin{tabular}[c]{@{}l@{}}VGG-11\\ (CIFAR-10)\end{tabular}}}} & \multicolumn{1}{l|}{\textbf{Orthogonal Learning (ours)}} & \textbf{0} & \textbf{14.3 / 31.7} & \textbf{8.6 / 22.0} & \textbf{8.0 / 21.3} & \textbf{13.8 / 34.1} \\ \hline
\multicolumn{1}{|l|}{} & \multicolumn{1}{l|}{No defense} & \textbf{0} & 78.7 / 81.3 & 100 / 100 & 100 / 100 & 100 / 100 \\ \cline{2-7} 
\multicolumn{1}{|l|}{} & \multicolumn{1}{l|}{JPEG-Quality=95} & 2.2 & 74.8 / 79.9 & 100 / 100 & 100 / 100 & 100 / 100 \\ \cline{2-7} 
\multicolumn{1}{|l|}{} & \multicolumn{1}{l|}{JPEG-Quality=90} & 3.7 & 71.1 / 77.6 & 98.2 / 100 & 97.1 / 99.5 & 99.8 / 100 \\ \cline{2-7} 
\multicolumn{1}{|l|}{} & \multicolumn{1}{l|}{JPEG-Quality=85} & 6.2 & 64.9 / 77.3 & 82.7 / 98.1 & 85.3 / 97.5 & 95.7 / 99.8 \\ \cline{2-7} 
\multicolumn{1}{|l|}{} & \multicolumn{1}{l|}{JPEG-Quality=80} & 9.1 & 57.2 / 75.7 & 63.4 / 89.3 & 68.1 / 93.8 & 82.1 / 98.6 \\ \cline{2-7} 
\multicolumn{1}{|l|}{} & \multicolumn{1}{l|}{TVM-Weight=3} & 1.1 & \textbf{40.1 / 60.0} & 51.0 / 79.9 & 52.0 / 83.8 & 99.9 / 100 \\ \cline{2-7} 
\multicolumn{1}{|l|}{} & \multicolumn{1}{l|}{TVM-Weight=5} & 4.4 & 74.4 / 78.9 & 93.4 / 97.8 & 89.9 / 97.8 & 98.4 / 99.9 \\ \cline{2-7} 
\multicolumn{1}{|l|}{} & \multicolumn{1}{l|}{TVM-Weight=7} & 5.4 & 71.0 / 76.7 & 86.2 / 92.6 & 79.8 / 90.5 & 95.3 / 99.7 \\ \cline{2-7} 
\multicolumn{1}{|l|}{} & \multicolumn{1}{l|}{TVM-Weight=9} & 7.6 & 66.2 / 74.0 & 78.6 / 87.3 & 68.3 / 82.9 & 91.1 / 98.8 \\ \cline{2-7} 
\multicolumn{1}{|l|}{} & \multicolumn{1}{l|}{Bit Reduction - Depth=1} & 71.8 & 75.3 / 78.6 & 74.2 / 76.9 & 75.9 / 77.4 & 75.8 / 79.8 \\ \cline{2-7} 
\multicolumn{1}{|l|}{} & \multicolumn{1}{l|}{Bit Reduction - Depth=2} & 42.5 & 63.1 / 73.1 & 68.3 / 80.8 & 71.4 / 87.4 & 79.4 / 94.5 \\ \cline{2-7} 
\multicolumn{1}{|l|}{} & \multicolumn{1}{l|}{Bit Reduction - Depth=3} & 12.8 & 71.0 / 78.3 & 96.0 / 99.9 & 98.5 / 99.8 & 99.5 / 99.9 \\ \cline{2-7} 
\multicolumn{1}{|l|}{} & \multicolumn{1}{l|}{Bit Reduction - Depth=4} & 3.6 & 77.2 / 81.3 & 99.9 / 100 & 99.9 / 100 & 100 / 100 \\ \cline{2-7} 
\multicolumn{1}{|l|}{} & \multicolumn{1}{l|}{Bilateral Filtering - Window=3} & 17.8 & 65.9 / 76.7 & 85.4 / 95.5 & 82.2 / 93.6 & 94.4 / 99.5 \\ \cline{2-7} 
\multicolumn{1}{|l|}{} & \multicolumn{1}{l|}{Bilateral Filtering - Window=5} & 29.5 & 59.7 / 70.4 & 62.8 / 72.8 & 62.5 / 78.1 & 74.7 / 89.0 \\ \cline{2-7} 
\multicolumn{1}{|l|}{} & \multicolumn{1}{l|}{Bilateral Filtering - Window=7} & 36.2 & 63.6 / 71.1 & 69.2 / 74.0 & 67.6 / 78.8 & 76.2 / 90.7 \\ \cline{2-7} 
\multicolumn{1}{|l|}{} & \multicolumn{1}{l|}{Bilateral Filtering - Window=9} & 37.1 & 63.6 / 71.5 & 68.1 / 74.3 & 67.3 / 76.1 & 76.8 / 88.1 \\ \cline{2-7} 
\multicolumn{1}{|l|}{\multirow{-18}{*}{\textbf{\begin{tabular}[c]{@{}l@{}}ResNet20\\ (CIFAR-10)\end{tabular}}}} & \multicolumn{1}{l|}{\textbf{Orthogonal Learning (ours)}} & \textbf{0} & 52.0 / 74.0 & \textbf{47.7 / 69.3} & \textbf{47.5 / 73.4} & \textbf{54.4 / 76.8} \\ \hline
\multicolumn{1}{|l|}{} & \multicolumn{1}{l|}{No defense} & 0 & 93.0 / 93.1 & 97.9 / 98.6 & 98.1 / 99.3 & 97.3 / 97.7 \\ \cline{2-7} 
\multicolumn{1}{|l|}{} & \multicolumn{1}{l|}{JPEG-Quality=95} & 2.3 & 92.6 / 92.7 & 97.9 / 98.6 & 98.1 / 99.3 & 97.3 / 97.7 \\ \cline{2-7} 
\multicolumn{1}{|l|}{} & \multicolumn{1}{l|}{JPEG-Quality=90} & 6.2 & 91.9 / 92.1 & 97.9 / 98.6 & 98.1 / 99.3 & 97.3 / 97.7 \\ \cline{2-7} 
\multicolumn{1}{|l|}{} & \multicolumn{1}{l|}{JPEG-Quality=85} & 10.0 & 91.7 / 91.5 & 97.9 / 98.6 & 98.0 / 99.3 & 97.3 / 97.7 \\ \cline{2-7} 
\multicolumn{1}{|l|}{} & \multicolumn{1}{l|}{JPEG-Quality=80} & 14.1 & 90.5 / 90.8 & 96.6 / 98.5 & 97.3 / 99.1 & 97.3 / 97.7 \\ \cline{2-7} 
\multicolumn{1}{|l|}{} & \multicolumn{1}{l|}{TVM-Weight=3} & 9.6 & 93.4 / 93.1 & 97.3 / 98.6 & 98.1 / 99.3 & 97.3 / 97.7 \\ \cline{2-7} 
\multicolumn{1}{|l|}{} & \multicolumn{1}{l|}{TVM-Weight=5} & 14.9 & 93.1 / 93.3 & 95.7 / 97.9 & 95.8 / 99.3 & 97.3 / 97.7 \\ \cline{2-7} 
\multicolumn{1}{|l|}{} & \multicolumn{1}{l|}{TVM-Weight=7} & 22.2 & 93.0 / 93.5 & 92.6 / 97.2 & 92.8 / 98.3 & 96.9 / 97.7 \\ \cline{2-7} 
\multicolumn{1}{|l|}{} & \multicolumn{1}{l|}{TVM-Weight=9} & 27.5 & 92.8 / 92.9 & 88.7 / 95.4 & 87.9 / 96.1 & 96.3 / 97.7 \\ \cline{2-7} 
\multicolumn{1}{|l|}{} & \multicolumn{1}{l|}{Bit Reduction - Depth=1} & 95.2 & 96.4 / 96.8 & 95.9 / 96.3 & 96.1 / 97.4 & 96.6 / 96.4 \\ \cline{2-7} 
\multicolumn{1}{|l|}{} & \multicolumn{1}{l|}{Bit Reduction - Depth=2} & 84.6 & 93.6 / 94.8 & 90.8 / 93.6 & 93.2 / 97.0 & 94.4 / 97.5 \\ \cline{2-7} 
\multicolumn{1}{|l|}{} & \multicolumn{1}{l|}{Bit Reduction - Depth=3} & 40.6 & 90.7 / 91.8 & 96.4 / 98.3 & 97.8 / 99.3 & 97.3 / 97.7 \\ \cline{2-7} 
\multicolumn{1}{|l|}{} & \multicolumn{1}{l|}{Bit Reduction - Depth=4} & 10.8 & 92.4 / 92.9 & 97.9 / 98.6 & 98.1 / 99.3 & 97.3 / 97.7 \\ \cline{2-7} 
\multicolumn{1}{|l|}{} & \multicolumn{1}{l|}{Bilateral Filtering - Window=3} & 18.1 & 90.1 / 90.9 & 97.6 / 98.1 & 96.3 / 98.7 & 97.2 / 97.7 \\ \cline{2-7} 
\multicolumn{1}{|l|}{} & \multicolumn{1}{l|}{Bilateral Filtering - Window=5} & 22.3 & 87.8 / 90.3 & 94.5 / 97.5 & 89.1 / 96.7 & 96.4 / 97.6 \\ \cline{2-7} 
\multicolumn{1}{|l|}{} & \multicolumn{1}{l|}{Bilateral Filtering - Window=7} & 19.3 & 87.7 / 90.1 & 96.0 / 97.7 & 90.8 / 97.0 & 96.5 / 97.6 \\ \cline{2-7} 
\multicolumn{1}{|l|}{} & \multicolumn{1}{l|}{Bilateral Filtering - Window=9} & 18.9 & 86.9 / 89.5 & 95.2 / 97.4 & 89.8 / 96.2 & 96.4 / 97.6 \\ \cline{2-7} 
\multicolumn{1}{|l|}{\multirow{-18}{*}{\textbf{\begin{tabular}[c]{@{}l@{}}VGG-16 \\ (ImageNet)\end{tabular}}}} & \multicolumn{1}{l|}{\textbf{Orthogonal Learning (ours)}} &  \textbf{0} & \textbf{37.2 / 61.8} & \textbf{14.0 / 30.8} & \textbf{19.4 / 47.8} & \textbf{32.3 / 61.4} \\ \hline
\end{tabular}%
}
\caption{\textcolor{black}{Comparison of fooling ratio (\%) of the proposed orthogonal learning with other defense schemes i.e.,~JPEG compression, Total Variation Minimization (TVM), Bit Depth Reduction (BR), Bilateral Filtering (BR). Different variants of defense schemes are generated by controlling the hyper-parameter as indicated in the second column. In the last four columns, the results are shown for the attacks generated for $\epsilon_1=0.03$ and $\epsilon_2=0.05$ in the order $\epsilon_1/\epsilon_2$ . The best results (smaller values) for each model and attack scheme are indicated in bold fonts.}}
\label{orthoVsOthersComparison}
\end{table*}

\subsection{Comparison with Other Techniques}
\cs{In the previous discussion, we illustrated the empirical results of our proposed defense strategy against different adversarial attacks. In this Section, we extend the results by a detailed comparison of our technique with related defense algorithms that are available in the literature. }

\cs
{We compare our technique with four other schemes, including JPEG compression \cite{guo2017countering}, Total Variance minimization \cite{rudin1994total}, Bit Squeezing \cite{xu2017feature} and Bilateral Filtering \cite{xie2019feature}.
We comprehensively evaluate existing defense strategies by using different settings of their hyperparameters in the ranges that are commonly used. Therefore, the results are reported for four different variants of each method, making our comparison considerably thorough. We refer to the original works for details on the significance of hyperparameters. 
% For each defense scheme, different variants {\color{red} (of what, not clear)} were created by careful selection of hyper-parameters such that the accuracy on clean samples is retained, which is also the quality of our technique. 
In JPEG compression, we sweep the `quality' hyper-parameter between  80$\%$ and 95$\%$, similarly `weights' are changed from $3$ to $9$ for the TVM and for bilateral filters the `window size' is varied between $3$ and $9$. For bit Squeezing defenses the `depth' is selected between $1$ and $4$. }

\cs{In Table \ref{orthoVsOthersComparison}, we compare the performance of defense algorithms for the models trained over ImageNet and CIFAR10 as discussed in Table \ref{simTableCIFAR10} and \ref{simTable}. The results are shown for $1000$ randomly sampled images from the validation dataset, such that the clean version of these images are correctly classified by the standard as well as the orthogonal model. 
The architecture under attack/defense is highlighted in the first column of Table \ref{orthoVsOthersComparison}.  
Perturbations are computed for the source model via four different attack schemes for $\epsilon_1=0.03$ and $\epsilon_2=0.05$. The last four columns report the fooling ratios in order $\epsilon_1/\epsilon_2$. The choice of $\epsilon$ values are based on the literature, such that a higher value is usually perceptible \cite{moosavi2016deepfool}. The 
raw fooling accuracy (no defense) is shown in the first row for each model.  These perturbations are then defended via different schemes and the results are reported in the subsequent rows. The best results for each attack type and model are indicated in bold. }

\cs{The results in Table \ref{orthoVsOthersComparison} validate the superior performance of our technique against others by a significant margin for all the attacks. Only in the case of FGSM attacks against ResNet20, our algorithm has shown inferior performance. This phenomena is inline with the earlier discussed performance of ResNet-20 model against the FGSM attacks in Figure \ref{robustnessFigCIFAR10}. Besides the improvement in adversarial robustness, it is important to notice the relative clean sample accuracy of the defense strategies. All other techniques suffer from a noticeable drop in the clean samples accuracy that limits the scope of their practical application. Our method maintains a zero fooling rate on clean samples for all models.}

\section{Conclusion}
We  introduced a model-agnostic gradient regularization scheme that allows to manifest orthogonal variants of a model. 
These orthogonal versions exhibit enhanced resistance to adversarial perturbations against black-box attacks in scenarios where the attacker is even able to guess the architecture of the target model. 
\cs{In such scenario the different architectures trained over CIFAR-10 show a minimum enhancement of $18.5\%$ against PGD, $9.1\%$ against I-FGSM and $17.4\%$ against the MI-FGSM attacks for $\epsilon=0.05$.}
Importantly, our regularization only results in a small degradation in the performance of the original models. This is in contrast to the conventional techniques for making the models robust to adversarial attacks, which generally result in significant loss of the original model  accuracy.  
We presented our results for  small-scale as well as large-scale datasets to establish the proposed technique. The achieved results highlight the potential of our scheme against the strongest available attacks.  
%However, a more detailed treatment involving over one million samples of the entire ImageNet and modern deep architectures like ResNet \cite{he2016deep} and DenseNets \cite{huang2017densely} is needed to unveil further potentials of self-orthogonal networks. 

\vspace{3mm}
\noindent\textbf{Acknowledgment} This  research  was  supported  by ARC Discovery Grant DP190102443, DP150100294 and DP150104251. The Titan V used in our experiments was donated by NVIDIA corporation.

%\newpage
\bibliographystyle{IEEEtran}
\bibliography{Explanation.bib}

% Generated by IEEEtran.bst, version: 1.14 (2015/08/26)
\begin{thebibliography}{10}
\providecommand{\url}[1]{#1}
\csname url@samestyle\endcsname
\providecommand{\newblock}{\relax}
\providecommand{\bibinfo}[2]{#2}
\providecommand{\BIBentrySTDinterwordspacing}{\spaceskip=0pt\relax}
\providecommand{\BIBentryALTinterwordstretchfactor}{4}
\providecommand{\BIBentryALTinterwordspacing}{\spaceskip=\fontdimen2\font plus
\BIBentryALTinterwordstretchfactor\fontdimen3\font minus
  \fontdimen4\font\relax}
\providecommand{\BIBforeignlanguage}[2]{{%
\expandafter\ifx\csname l@#1\endcsname\relax
\typeout{** WARNING: IEEEtran.bst: No hyphenation pattern has been}%
\typeout{** loaded for the language `#1'. Using the pattern for}%
\typeout{** the default language instead.}%
\else
\language=\csname l@#1\endcsname
\fi
#2}}
\providecommand{\BIBdecl}{\relax}
\BIBdecl

\bibitem{krizhevsky2012imagenet}
A.~Krizhevsky, I.~Sutskever, and G.~E. Hinton, ``Imagenet classification with
  deep convolutional neural networks,'' in \emph{Advances in neural information
  processing systems}, 2012, pp. 1097--1105.

\bibitem{simonyan2014very}
K.~Simonyan and A.~Zisserman, ``Very deep convolutional networks for
  large-scale image recognition,'' \emph{arXiv preprint arXiv:1409.1556}, 2014.

\bibitem{redmon2017yolo9000}
J.~Redmon and A.~Farhadi, ``Yolo9000: better, faster, stronger,'' in
  \emph{Proceedings of the IEEE conference on computer vision and pattern
  recognition}, 2017, pp. 7263--7271.

\bibitem{ren2015faster}
S.~Ren, K.~He, R.~Girshick, and J.~Sun, ``Faster r-cnn: Towards real-time
  object detection with region proposal networks,'' in \emph{Advances in neural
  information processing systems}, 2015, pp. 91--99.

\bibitem{long2015fully}
J.~Long, E.~Shelhamer, and T.~Darrell, ``Fully convolutional networks for
  semantic segmentation,'' in \emph{Proceedings of the IEEE conference on
  computer vision and pattern recognition}, 2015, pp. 3431--3440.

\bibitem{chen2018encoder}
L.-C. Chen, Y.~Zhu, G.~Papandreou, F.~Schroff, and H.~Adam, ``Encoder-decoder
  with atrous separable convolution for semantic image segmentation,'' in
  \emph{Proceedings of the European conference on computer vision (ECCV)},
  2018, pp. 801--818.

\bibitem{wu2017image}
Q.~Wu, C.~Shen, P.~Wang, A.~Dick, and A.~van~den Hengel, ``Image captioning and
  visual question answering based on attributes and external knowledge,''
  \emph{IEEE transactions on pattern analysis and machine intelligence},
  vol.~40, no.~6, pp. 1367--1381, 2017.

\bibitem{szegedy2013intriguing}
C.~Szegedy, W.~Zaremba, I.~Sutskever, J.~Bruna, D.~Erhan, I.~Goodfellow, and
  R.~Fergus, ``Intriguing properties of neural networks,'' \emph{arXiv preprint
  arXiv:1312.6199}, 2013.

\bibitem{akhtar2018threat}
N.~Akhtar and A.~Mian, ``Threat of adversarial attacks on deep learning in
  computer vision: A survey,'' \emph{IEEE Access}, vol.~6, pp.
  14\,410--14\,430, 2018.

\bibitem{andriushchenko2019square}
M.~Andriushchenko, F.~Croce, N.~Flammarion, and M.~Hein, ``Square attack: a
  query-efficient black-box adversarial attack via random search,'' \emph{arXiv
  preprint arXiv:1912.00049}, 2019.

\bibitem{chen2019hopskipjumpattack}
J.~Chen, M.~I. Jordan, and M.~J. Wainwright, ``Hopskipjumpattack: A
  query-efficient decision-based attack,'' \emph{arXiv preprint
  arXiv:1904.02144}, vol.~3, 2019.

\bibitem{ilyas2018black}
A.~Ilyas, L.~Engstrom, A.~Athalye, and J.~Lin, ``Black-box adversarial attacks
  with limited queries and information,'' \emph{arXiv preprint
  arXiv:1804.08598}, 2018.

\bibitem{madry2017towards}
A.~Madry, A.~Makelov, L.~Schmidt, D.~Tsipras, and A.~Vladu, ``Towards deep
  learning models resistant to adversarial attacks,'' \emph{arXiv preprint
  arXiv:1706.06083}, 2017.

\bibitem{liu2016delving}
Y.~Liu, X.~Chen, C.~Liu, and D.~Song, ``Delving into transferable adversarial
  examples and black-box attacks,'' \emph{arXiv preprint arXiv:1611.02770},
  2016.

\bibitem{dong2018boosting}
Y.~Dong, F.~Liao, T.~Pang, H.~Su, J.~Zhu, X.~Hu, and J.~Li, ``Boosting
  adversarial attacks with momentum,'' in \emph{Proceedings of the IEEE
  conference on computer vision and pattern recognition}, 2018, pp. 9185--9193.

\bibitem{xie2019improving}
C.~Xie, Z.~Zhang, Y.~Zhou, S.~Bai, J.~Wang, Z.~Ren, and A.~L. Yuille,
  ``Improving transferability of adversarial examples with input diversity,''
  in \emph{Proceedings of the IEEE Conference on Computer Vision and Pattern
  Recognition}, 2019, pp. 2730--2739.

\bibitem{goodfellow2014explaining}
I.~J. Goodfellow, J.~Shlens, and C.~Szegedy, ``Explaining and harnessing
  adversarial examples,'' \emph{arXiv preprint arXiv:1412.6572}, 2014.

\bibitem{kurakin2016adversarial}
A.~Kurakin, I.~Goodfellow, and S.~Bengio, ``Adversarial examples in the
  physical world,'' \emph{arXiv preprint arXiv:1607.02533}, 2016.

\bibitem{wu2018understanding}
L.~Wu, Z.~Zhu, C.~Tai \emph{et~al.}, ``Understanding and enhancing the
  transferability of adversarial examples,'' \emph{arXiv preprint
  arXiv:1802.09707}, 2018.

\bibitem{moosavi2016deepfool}
S.-M. Moosavi-Dezfooli, A.~Fawzi, and P.~Frossard, ``Deepfool: a simple and
  accurate method to fool deep neural networks,'' in \emph{Proceedings of the
  IEEE conference on computer vision and pattern recognition}, 2016, pp.
  2574--2582.

\bibitem{shi2019curls}
Y.~Shi, S.~Wang, and Y.~Han, ``Curls \& whey: Boosting black-box adversarial
  attacks,'' \emph{arXiv preprint arXiv:1904.01160}, 2019.

\bibitem{rony2019decoupling}
J.~Rony, L.~G. Hafemann, L.~S. Oliveira, I.~B. Ayed, R.~Sabourin, and
  E.~Granger, ``Decoupling direction and norm for efficient gradient-based l2
  adversarial attacks and defenses,'' in \emph{Proceedings of the IEEE
  Conference on Computer Vision and Pattern Recognition}, 2019, pp. 4322--4330.

\bibitem{croce2019sparse}
F.~Croce and M.~Hein, ``Sparse and imperceivable adversarial attacks,''
  \emph{arXiv preprint arXiv:1909.05040}, 2019.

\bibitem{zheng2019distributionally}
T.~Zheng, C.~Chen, and K.~Ren, ``Distributionally adversarial attack,'' in
  \emph{Proceedings of the AAAI Conference on Artificial Intelligence},
  vol.~33, 2019, pp. 2253--2260.

\bibitem{ganeshan2019fda}
A.~Ganeshan and R.~V. Babu, ``Fda: Feature disruptive attack,'' in
  \emph{Proceedings of the IEEE International Conference on Computer Vision},
  2019, pp. 8069--8079.

\bibitem{moosavi2017universal}
S.-M. Moosavi-Dezfooli, A.~Fawzi, O.~Fawzi, and P.~Frossard, ``Universal
  adversarial perturbations,'' in \emph{Proceedings of the IEEE conference on
  computer vision and pattern recognition}, 2017, pp. 1765--1773.

\bibitem{akhtar2019label}
N.~Akhtar, M.~A. Jalwana, M.~Bennamoun, and A.~Mian, ``Label universal targeted
  attack,'' \emph{arXiv preprint arXiv:1905.11544}, 2019.

\bibitem{papernot2017practical}
N.~Papernot, P.~McDaniel, I.~Goodfellow, S.~Jha, Z.~B. Celik, and A.~Swami,
  ``Practical black-box attacks against machine learning,'' in
  \emph{Proceedings of the 2017 ACM on Asia conference on computer and
  communications security}, 2017, pp. 506--519.

\bibitem{bhambri2019survey}
S.~Bhambri, S.~Muku, A.~Tulasi, and A.~Balaji~Buduru, ``A survey of black-box
  adversarial attacks on computer vision models,'' \emph{arXiv}, pp.
  arXiv--1912, 2019.

\bibitem{dong2019evading}
Y.~Dong, T.~Pang, H.~Su, and J.~Zhu, ``Evading defenses to transferable
  adversarial examples by translation-invariant attacks,'' in \emph{Proceedings
  of the IEEE Conference on Computer Vision and Pattern Recognition}, 2019, pp.
  4312--4321.

\bibitem{ilyas2019adversarial}
A.~Ilyas, S.~Santurkar, D.~Tsipras, L.~Engstrom, B.~Tran, and A.~Madry,
  ``Adversarial examples are not bugs, they are features,'' \emph{arXiv
  preprint arXiv:1905.02175}, 2019.

\bibitem{jia2019comdefend}
X.~Jia, X.~Wei, X.~Cao, and H.~Foroosh, ``Comdefend: An efficient image
  compression model to defend adversarial examples,'' in \emph{Proceedings of
  the IEEE Conference on Computer Vision and Pattern Recognition}, 2019, pp.
  6084--6092.

\bibitem{akhtar2018defense}
N.~Akhtar, J.~Liu, and A.~Mian, ``Defense against universal adversarial
  perturbations,'' in \emph{Proceedings of the IEEE Conference on Computer
  Vision and Pattern Recognition}, 2018, pp. 3389--3398.

\bibitem{liu2019detection}
J.~Liu, W.~Zhang, Y.~Zhang, D.~Hou, Y.~Liu, H.~Zha, and N.~Yu, ``Detection
  based defense against adversarial examples from the steganalysis point of
  view,'' in \emph{Proceedings of the IEEE Conference on Computer Vision and
  Pattern Recognition}, 2019, pp. 4825--4834.

\bibitem{miyato2016adversarial}
T.~Miyato, A.~M. Dai, and I.~Goodfellow, ``Adversarial training methods for
  semi-supervised text classification,'' \emph{arXiv preprint
  arXiv:1605.07725}, 2016.

\bibitem{woods2019reliable}
W.~Woods, J.~Chen, and C.~Teuscher, ``Reliable classification explanations via
  adversarial attacks on robust networks,'' \emph{arXiv preprint
  arXiv:1906.02896}, 2019.

\bibitem{gu2014towards}
S.~Gu and L.~Rigazio, ``Towards deep neural network architectures robust to
  adversarial examples,'' \emph{arXiv preprint arXiv:1412.5068}, 2014.

\bibitem{bai2017alleviating}
W.~Bai, C.~Quan, and Z.~Luo, ``Alleviating adversarial attacks via
  convolutional autoencoder,'' in \emph{2017 18th IEEE/ACIS International
  Conference on Software Engineering, Artificial Intelligence, Networking and
  Parallel/Distributed Computing (SNPD)}.\hskip 1em plus 0.5em minus
  0.4em\relax IEEE, 2017, pp. 53--58.

\bibitem{gao2017deepcloak}
J.~Gao, B.~Wang, Z.~Lin, W.~Xu, and Y.~Qi, ``Deepcloak: Masking deep neural
  network models for robustness against adversarial samples,'' \emph{arXiv
  preprint arXiv:1702.06763}, 2017.

\bibitem{ross2018improving}
A.~S. Ross and F.~Doshi-Velez, ``Improving the adversarial robustness and
  interpretability of deep neural networks by regularizing their input
  gradients,'' in \emph{Thirty-second AAAI conference on artificial
  intelligence}, 2018.

\bibitem{lyu2015unified}
C.~Lyu, K.~Huang, and H.-N. Liang, ``A unified gradient regularization family
  for adversarial examples,'' in \emph{2015 IEEE International Conference on
  Data Mining}.\hskip 1em plus 0.5em minus 0.4em\relax IEEE, 2015, pp.
  301--309.

\bibitem{shaham2018understanding}
U.~Shaham, Y.~Yamada, and S.~Negahban, ``Understanding adversarial training:
  Increasing local stability of supervised models through robust
  optimization,'' \emph{Neurocomputing}, vol. 307, pp. 195--204, 2018.

\bibitem{shen2017ape}
S.~Shen, G.~Jin, K.~Gao, and Y.~Zhang, ``Ape-gan: Adversarial perturbation
  elimination with gan,'' \emph{arXiv preprint arXiv:1707.05474}, 2017.

\bibitem{lee2017generative}
H.~Lee, S.~Han, and J.~Lee, ``Generative adversarial trainer: Defense to
  adversarial perturbations with gan,'' \emph{arXiv preprint arXiv:1705.03387},
  2017.

\bibitem{prakash2018deflecting}
A.~Prakash, N.~Moran, S.~Garber, A.~DiLillo, and J.~Storer, ``Deflecting
  adversarial attacks with pixel deflection,'' in \emph{Proceedings of the IEEE
  conference on computer vision and pattern recognition}, 2018, pp. 8571--8580.

\bibitem{raff2019barrage}
E.~Raff, J.~Sylvester, S.~Forsyth, and M.~McLean, ``Barrage of random
  transforms for adversarially robust defense,'' in \emph{Proceedings of the
  IEEE Conference on Computer Vision and Pattern Recognition}, 2019, pp.
  6528--6537.

\bibitem{guo2017countering}
C.~Guo, M.~Rana, M.~Cisse, and L.~Van Der~Maaten, ``Countering adversarial
  images using input transformations,'' \emph{arXiv preprint arXiv:1711.00117},
  2017.

\bibitem{carlini2017adversarial}
N.~Carlini and D.~Wagner, ``Adversarial examples are not easily detected:
  Bypassing ten detection methods,'' in \emph{Proceedings of the 10th ACM
  Workshop on Artificial Intelligence and Security}.\hskip 1em plus 0.5em minus
  0.4em\relax ACM, 2017, pp. 3--14.

\bibitem{carlini2016defensive}
------, ``Defensive distillation is not robust to adversarial examples,''
  \emph{arXiv preprint arXiv:1607.04311}, 2016.

\bibitem{athalye2018robustness}
A.~Athalye and N.~Carlini, ``On the robustness of the cvpr 2018 white-box
  adversarial example defenses,'' \emph{arXiv preprint arXiv:1804.03286}, 2018.

\bibitem{santurkar2019computer}
S.~Santurkar, D.~Tsipras, B.~Tran, A.~Ilyas, L.~Engstrom, and A.~Madry,
  ``Computer vision with a single (robust) classifier,'' \emph{arXiv preprint
  arXiv:1906.09453}, 2019.

\bibitem{ourcvpr2020}
M.~A. Jalwana, N.~Akhtar, M.~Bennamoun, and A.~Mian, ``Attack to explain deep
  representation,'' in \emph{Proceedings of the IEEE Conference on Computer
  Vision and Pattern Recognition}, 2020.

\bibitem{zhang2018cappronet}
L.~Zhang, M.~Edraki, and G.-J. Qi, ``Cappronet: Deep feature learning via
  orthogonal projections onto capsule subspaces,'' in \emph{Advances in Neural
  Information Processing Systems}, 2018, pp. 5814--5823.

\bibitem{gayer2020convolutional}
A.~V. Gayer and A.~V. Sheshkus, ``Convolutional neural network weights
  regularization via orthogonalization,'' in \emph{Twelfth International
  Conference on Machine Vision (ICMV 2019)}, vol. 11433.\hskip 1em plus 0.5em
  minus 0.4em\relax International Society for Optics and Photonics, 2020, p.
  1143326.

\bibitem{jia2019orthogonal}
K.~Jia, S.~Li, Y.~Wen, T.~Liu, and D.~Tao, ``Orthogonal deep neural networks,''
  \emph{IEEE transactions on pattern analysis and machine intelligence}, 2019.

\bibitem{kariyappa2019improving}
S.~Kariyappa and M.~K. Qureshi, ``Improving adversarial robustness of ensembles
  with diversity training,'' \emph{arXiv preprint arXiv:1901.09981}, 2019.

\bibitem{gorban2018blessing}
A.~N. Gorban and I.~Y. Tyukin, ``Blessing of dimensionality: mathematical
  foundations of the statistical physics of data,'' \emph{Philosophical
  Transactions of the Royal Society A: Mathematical, Physical and Engineering
  Sciences}, vol. 376, no. 2118, p. 20170237, 2018.

\bibitem{selvaraju2017grad}
R.~R. Selvaraju, M.~Cogswell, A.~Das, R.~Vedantam, D.~Parikh, and D.~Batra,
  ``Grad-cam: Visual explanations from deep networks via gradient-based
  localization,'' in \emph{Proceedings of the IEEE International Conference on
  Computer Vision}, 2017, pp. 618--626.

\bibitem{adebayo2018sanity}
J.~Adebayo, J.~Gilmer, M.~Muelly, I.~Goodfellow, M.~Hardt, and B.~Kim, ``Sanity
  checks for saliency maps,'' in \emph{Advances in Neural Information
  Processing Systems}, 2018, pp. 9505--9515.

\bibitem{krizhevsky2009learning}
A.~Krizhevsky, G.~Hinton \emph{et~al.}, ``Learning multiple layers of features
  from tiny images,'' 2009.

\bibitem{deng2009imagenet}
J.~Deng, W.~Dong, R.~Socher, L.-J. Li, K.~Li, and L.~Fei-Fei, ``Imagenet: A
  large-scale hierarchical image database,'' in \emph{2009 IEEE conference on
  computer vision and pattern recognition}.\hskip 1em plus 0.5em minus
  0.4em\relax Ieee, 2009, pp. 248--255.

\bibitem{rauber2017foolbox}
J.~Rauber, W.~Brendel, and M.~Bethge, ``Foolbox: a python toolbox to benchmark
  the robustness of machine learning models (2017),'' \emph{URL http://arxiv.
  org/abs/1707.04131}, vol.~6, 2017.

\bibitem{ding2019advertorch}
G.~W. Ding, L.~Wang, and X.~Jin, ``{AdverTorch} v0.1: An adversarial robustness
  toolbox based on pytorch,'' \emph{arXiv preprint arXiv:1902.07623}, 2019.

\bibitem{NEURIPS2019_9015}
\BIBentryALTinterwordspacing
A.~Paszke, S.~Gross, F.~Massa, A.~Lerer, J.~Bradbury, G.~Chanan, T.~Killeen,
  Z.~Lin, N.~Gimelshein, L.~Antiga, A.~Desmaison, A.~Kopf, E.~Yang, Z.~DeVito,
  M.~Raison, A.~Tejani, S.~Chilamkurthy, B.~Steiner, L.~Fang, J.~Bai, and
  S.~Chintala, ``Pytorch: An imperative style, high-performance deep learning
  library,'' in \emph{Advances in Neural Information Processing Systems 32},
  H.~Wallach, H.~Larochelle, A.~Beygelzimer, F.~d\textquotesingle
  Alch\'{e}-Buc, E.~Fox, and R.~Garnett, Eds.\hskip 1em plus 0.5em minus
  0.4em\relax Curran Associates, Inc., 2019, pp. 8024--8035. [Online].
  Available:
  \url{http://papers.neurips.cc/paper/9015-pytorch-an-imperative-style-high-performance-deep-learning-library.pdf}
\BIBentrySTDinterwordspacing

\bibitem{chengyangfu}
C.-Y. Fu, ``Pytorch implementation of vgg network trained on cifar10 dataset,''
  \url{https://github.com/chengyangfu/pytorch-vgg-cifar10}, 2019.

\bibitem{akamaster}
Y.~Idelbayev, ``Resnets for cifar10/100 in pytorch,''
  \url{https://github.com/akamaster/pytorch_resnet_cifar10}, 2019.

\bibitem{wu2020skip}
D.~Wu, Y.~Wang, S.-T. Xia, J.~Bailey, and X.~Ma, ``Skip connections matter: On
  the transferability of adversarial examples generated with resnets,''
  \emph{arXiv preprint arXiv:2002.05990}, 2020.

\bibitem{glorot2010understanding}
X.~Glorot and Y.~Bengio, ``Understanding the difficulty of training deep
  feedforward neural networks,'' in \emph{Proceedings of the thirteenth
  international conference on artificial intelligence and statistics}, 2010,
  pp. 249--256.

\bibitem{lecun-mnisthandwrittendigit-2010}
\BIBentryALTinterwordspacing
Y.~LeCun and C.~Cortes, ``{MNIST} handwritten digit database,'' 2010. [Online].
  Available: \url{http://yann.lecun.com/exdb/mnist/}
\BIBentrySTDinterwordspacing

\bibitem{simon2018adversarial}
C.-J. Simon-Gabriel, Y.~Ollivier, L.~Bottou, B.~Sch{\"o}lkopf, and
  D.~Lopez-Paz, ``Adversarial vulnerability of neural networks increases with
  input dimension,'' \emph{arXiv preprint arXiv:1802.01421}, 2018.

\bibitem{he2016deep}
K.~He, X.~Zhang, S.~Ren, and J.~Sun, ``Deep residual learning for image
  recognition,'' in \emph{Proceedings of the IEEE conference on computer vision
  and pattern recognition}, 2016, pp. 770--778.

\bibitem{huang2017densely}
G.~Huang, Z.~Liu, L.~Van Der~Maaten, and K.~Q. Weinberger, ``Densely connected
  convolutional networks,'' in \emph{Proceedings of the IEEE conference on
  computer vision and pattern recognition}, 2017, pp. 4700--4708.

\bibitem{iandola2016squeezenet}
F.~N. Iandola, S.~Han, M.~W. Moskewicz, K.~Ashraf, W.~J. Dally, and K.~Keutzer,
  ``Squeezenet: Alexnet-level accuracy with 50x fewer parameters and< 0.5 mb
  model size,'' \emph{arXiv preprint arXiv:1602.07360}, 2016.

\bibitem{rudin1994total}
L.~I. Rudin and S.~Osher, ``Total variation based image restoration with free
  local constraints,'' in \emph{Proceedings of 1st International Conference on
  Image Processing}, vol.~1.\hskip 1em plus 0.5em minus 0.4em\relax IEEE, 1994,
  pp. 31--35.

\bibitem{xu2017feature}
W.~Xu, D.~Evans, and Y.~Qi, ``Feature squeezing: Detecting adversarial examples
  in deep neural networks,'' \emph{arXiv preprint arXiv:1704.01155}, 2017.

\bibitem{xie2019feature}
C.~Xie, Y.~Wu, L.~v.~d. Maaten, A.~L. Yuille, and K.~He, ``Feature denoising
  for improving adversarial robustness,'' in \emph{Proceedings of the IEEE
  Conference on Computer Vision and Pattern Recognition}, 2019, pp. 501--509.

\end{thebibliography}

\begin{IEEEbiography}[{\includegraphics[width=1in,height=1.25in,clip,keepaspectratio]{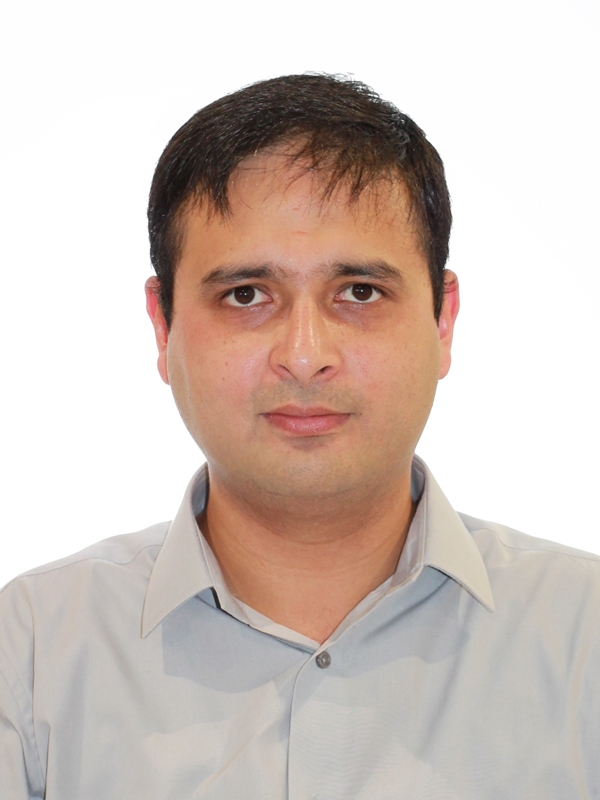}}]{Mohammad A. A. K. Jalwana} is currently working towards the PhD degree  at  The  University  of  Western  Australia (UWA) under the supervision of Prof. Ajmal Mian, Prof. Mohammed Bennamoun and  Dr.  Naveed  Akhtar.  He  is  a  recipient  of scholarship for international research fees (SIRF). His research interests are adversarial deep  learning  and  its  application in  computer vision, 3D localization, image captioning and medical image processing.
\end{IEEEbiography}

\begin{IEEEbiography}[{\includegraphics[width=1in,height=1.25in,clip,keepaspectratio]{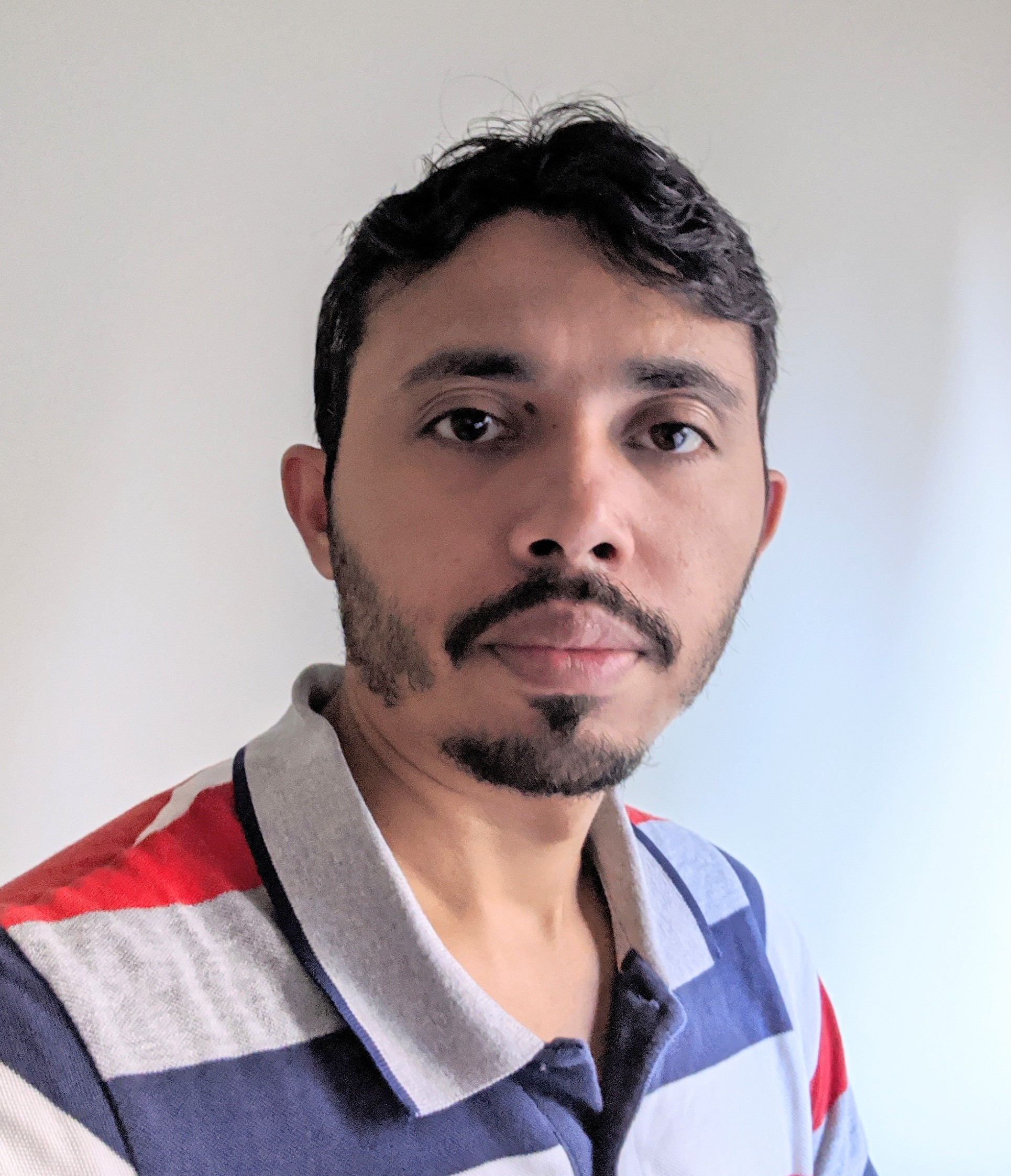}}]{Naveed Akhtar} received his PhD in Computer Vision from The University of Western Australia (UWA) and Master degree in Computer Science from Hochschule Bonn-Rhein-Sieg, Germany. His research in Computer Vision is regularly published in  prestigious sources of the filed. He also serves as an Associate Editor of IEEE Access.  
%During his PhD, he was  recipient of multiple scholarships, and  runner-up for the Canon Extreme Imaging Competition in 2015. 
Currently, he is a Lecturer at UWA. Previously, he also served as a Research Fellow at UWA and the Australian National University. His research interests include adversarial deep learning, multiple object tracking, action recognition,  and hyperspectral image analysis.
\end{IEEEbiography}

\begin{IEEEbiography}[{\includegraphics[width=1in,height=1.25in,clip]{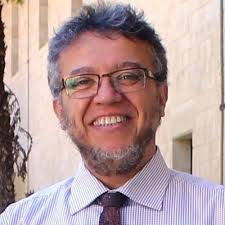}}]{MOHAMMED BENNAMOUN} is Winthrop Professor in the Department of Computer Science and Software Engineering at UWA and is a researcher in computer vision, machine/deep learning, robotics, and signal/speech processing. He has published 4 books (available on Amazon, 1 edited book, 1 Encyclopedia article, 14 book chapters, 140+ journal papers, 260+ conference publications, 16 invited $\&$ keynote publications. His h-index is 55 and his number of citations is 13,000+ (Google Scholar). He was awarded 65+ competitive research grants, from the Australian Research Council, and numerous other Government, UWA and industry Research Grants. He successfully supervised 26+ PhD students to completion. He won the Best Supervisor of the Year Award at QUT (1998), and received award for research supervision at UWA (2008 $\&$ 2016) and Vice-Chancellor Award for mentorship (2016).  He delivered conference tutorials at major conferences, including: IEEE Computer Vision and Pattern Recognition (CVPR 2016), Interspeech 2014, IEEE International Conference on Acoustics Speech and Signal Processing (ICASSP) and European Conference on Computer Vision (ECCV). He was also invited to give a Tutorial at an International Summer School on Deep Learning (DeepLearn 2017).
\end{IEEEbiography}

\begin{IEEEbiography}[{\includegraphics[width=1in,height=1.25in,clip,keepaspectratio]{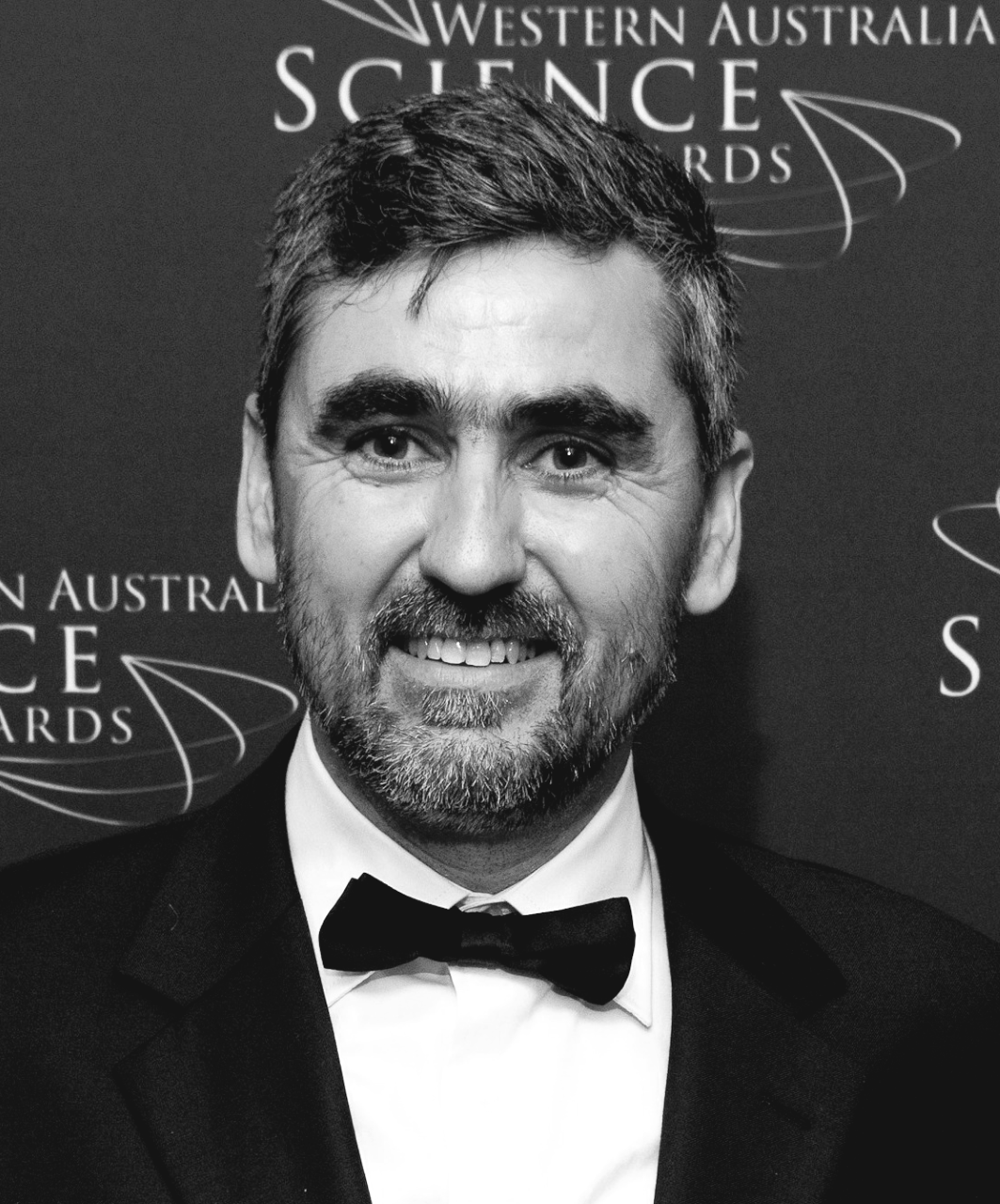}}]{Ajmal Mian} is a Professor of Computer Science at The University of Western Australia. He is the recipient of two prestigious national fellowships from the Australian Research Council and several awards including the West Australian Early  Career  Scientist  of  the  Year 2012, Excellence in Research Supervision, EH Thompson Award, ASPIRE Professional Development Award,  Vice-chancellors  Mid-career  Award,  Outstanding Young Investigator Award, the Australasian Distinguished Dissertation Award  and  various  best  paper  awards. He is an Associate Editor of IEEE Transactions on Neural Networks \& Learning Systems, IEEE Transactions on Image Processing and the Pattern Recognition journal. He served as the General Chair for DICTA 2019 and ACCV 2018. His  research  interests  are  in computer  vision,  deep  learning,  shape  analysis,  face  recognition, human action recognition and video analysis.
\end{IEEEbiography}

% You can push biographies down or up by placing
% a \vfill before or after them. The appropriate
% use of \vfill depends on what kind of text is
% on the last page and whether or not the columns
% are being equalized.

%\vfill

% Can be used to pull up biographies so that the bottom of the last one
% is flush with the other column.
%\enlargethispage{-5in}

% that's all folks
\end{document}